\documentclass[10pt,journal,compsoc]{IEEEtran}

\usepackage{graphicx}
\usepackage{balance}  
\usepackage{subfigure}

\usepackage{algorithm}
\usepackage{algorithmic}
\usepackage{amssymb}
\usepackage{amsmath}
\usepackage{hyperref}
\usepackage{macro}
\usepackage{amsthm}

\newtheorem{theorem}{Theorem}[section]
\newtheorem{lemma}{Lemma}[section]
\newtheorem{assumption}{Assumption}[section]

\ifCLASSOPTIONcompsoc
  \usepackage[nocompress]{cite}
\else
  \usepackage{cite}
\fi

\begin{document}

\title{Stochastic Gradient Made Stable: \\ A Manifold Propagation Approach for Large-Scale Optimization}

\author{Yadong~Mu,~\IEEEmembership{Member,~IEEE,}
        Wei~Liu,~\IEEEmembership{Member,~IEEE,}
        and~Wei~Fan,~\IEEEmembership{Member,~IEEE}
\IEEEcompsocitemizethanks{
\IEEEcompsocthanksitem Yadong Mu is a senior scientist of AT\&T Labs Research, Middletown, NJ 07748.\protect\\
E-mail: myd@research.att.com
\IEEEcompsocthanksitem Wei Liu is a Research Staff Member of IBM T. J. Watson Research Center, Yorktown Heights, NY 10598.\protect\\
E-mail: weiliu@us.ibm.com
\IEEEcompsocthanksitem Wei Fan is Director and Deputy Head of Baidu Big Data Research Lab, Sunny Vale, CA.\protect\\
E-mail: fanwei03@baidu.com
}
}

\IEEEtitleabstractindextext{%
\begin{abstract}
Stochastic gradient descent (SGD) holds as a classical method to build large scale machine learning models over big data. A stochastic gradient is typically calculated from a limited number of samples (known as mini-batch), which potentially incurs a high variance and causes the estimated parameters bounce around the optimal solution. To improve the stability of stochastic gradient, recent years have witnessed the proposal of several semi-stochastic gradient descent algorithms, which distinguish themselves from standard SGD by incorporating global information into gradient computation. In this paper we contribute a novel stratified semi-stochastic gradient descent (S3GD) algorithm to this nascent research area, accelerating the optimization of a large family of composite convex functions. Though theoretically converging faster, prior semi-stochastic algorithms are found to suffer from high iteration complexity, which makes them even slower than SGD in practice on many datasets. In our proposed S3GD, the semi-stochastic gradient is calculated based on efficient manifold propagation, which can be numerically accomplished by sparse matrix multiplications. This way S3GD is able to generate a highly-accurate estimate of the exact gradient from each mini-batch with largely-reduced computational complexity. Theoretic analysis reveals that the proposed S3GD elegantly balances the geometric algorithmic convergence rate against the space and time complexities during the optimization. The efficacy of S3GD is also experimentally corroborated on several large-scale benchmark datasets.
\end{abstract}

}

\maketitle

\IEEEdisplaynontitleabstractindextext
\IEEEpeerreviewmaketitle

\section{Introduction}
\label{sec:inro}

\emph{Regularized risk minimization}~\cite{Vapnik} is a fundamental subject in machine learning and statistics, whose formulations typically admit a combination of a loss function and a regularization term. This paper addresses a general class of convex regularized risk minimization problems which can be expressed as a composition:
\begin{eqnarray}
\w^\ast = \arg \min_{\w} ~ \{ F(\w) :=  P(\w^\top \x) + R(\w) \}, 
\label{eqn:com}
\end{eqnarray}
in which $\w,\x$ denote the parameter vector and data vector respectively. Both $P(\w^\top \x)$ and $R(\w)$ are assumed to be convex functions. Moreover, let $P(\w^\top \x)$ be a weighted addition of many atomic loss functions, each of which is differentiable. We simply define each atomic function on an input data pair $(\x_i,y_i)$, where $\x_i \in \mathbb{R}^d$ represents a feature vector and $y_i$ denotes its associated label. Popular choices of the loss functions include the square loss $(\w^\top \x_i-y_i)^2$, the logistic loss $\log (1+\exp(-y _i \w^\top \x_i))$, and the hinge loss $(1 - y_i \w^\top \x_i)_+$. In the above cases $y_i \in \{\pm 1\}$, yet in others $y_i$ can be real-valued in regression problems or missing in an unsupervised learning setting. $R(\w)$ defines a proper regularization function. It imposes some structural preference on the parameters (\emph{e.g.}, structural sparsity or matrix low-rankness). $R(\w)$ can be non-smooth with respect to $\w$, such as the sparsity-encouraging 1-norm $\|\w\|_1$.

When facing a large volume of training data, the space and time complexities become critical limiting factors in building a machine learning model. In such scenarios, stochastic (sub)gradient descent (SGD)~\cite{bottou-mlss-2004,Shalev-Shwartz013,abs-1012-1367,muyang,LiZCS14,YunYHVD14,GemullaNHS11} is a favored method used by many theorist and practitioners. The most attractive trait of SGD is the light-weight computation at each iteration of update. Its single-sample or mini-batch~\cite{CotterSSS11,Shalev-Shwartz013} updating scheme is a general remedy for the $\mathcal{O}(n)$ complexity in exact gradient descent (GD) methods ($n$ represents the number of training samples). Therefore, SGD algorithms are particularly promising when there is a limited budget of resources. Given properly-specified step size parameters at each iteration, SGD algorithms often enjoy provably fast rates of convergence.

The major downside of SGD in practical implementations is caused by large variance of stochastic gradients. Statistically, the mathematical expectation of stochastic gradients is exactly the full gradient. However, the randomness in constructing mini-batch brings large variance to stochastic gradients, particularly for complex data set. Moving along the direction of a stochastic gradient does not always guarantee a decrease of the entire training loss. Under large stochastic gradient variance, the estimated parameters often drastically bounce around the global optimal solution.

Recent years have witnessed the emerging efforts of developing sophisticated algorithms which reduce the stochastic gradient variance in SGD. The shared idea underlying these works is incorporating an additional gradient-correcting operation when computing the stochastic gradient. The corrected stochastic gradient becomes a more accurate approximation of the full gradient. Statistically, it enjoys a reduced level of variance. For example, the work in~\cite{ZhaoZ14} explicitly expresses the stochastic gradient variance and proves that constructing mini-batch using special non-uniform sampling strategy is able to reduce the stochastic gradient variance. The sampling probability is essentially based on the contextual importance of a sample. Another method named stochastic average gradient (SAG)~\cite{SchmidtRB13} keeps a record of historic stochastic gradients and adaptively averages them for the use in the current iteration. The rate of convergence is thereby improved to $\mathcal{O}(1/k)$ for general convex functions, and $\mathcal{O}(p^k)$ with $p < 1$ for strongly convex functions, respectively ($k$ is the count of iterations). For atomic functions in special forms (\emph{e.g.}, linear function of the data vectors as in linear regression and logistic regression), the storage of historic gradients in SAG can be reduced from $\mathcal{O}(n d)$ to $\mathcal{O}(n)$ ($n,d$ represent the sample count and feature dimension respectively). However, generally storing historic gradients in SAG entails a heavy burden for machine learning models with many parameters.

\begin{figure}[t]
\begin{center}
   \includegraphics[width=0.9\linewidth]{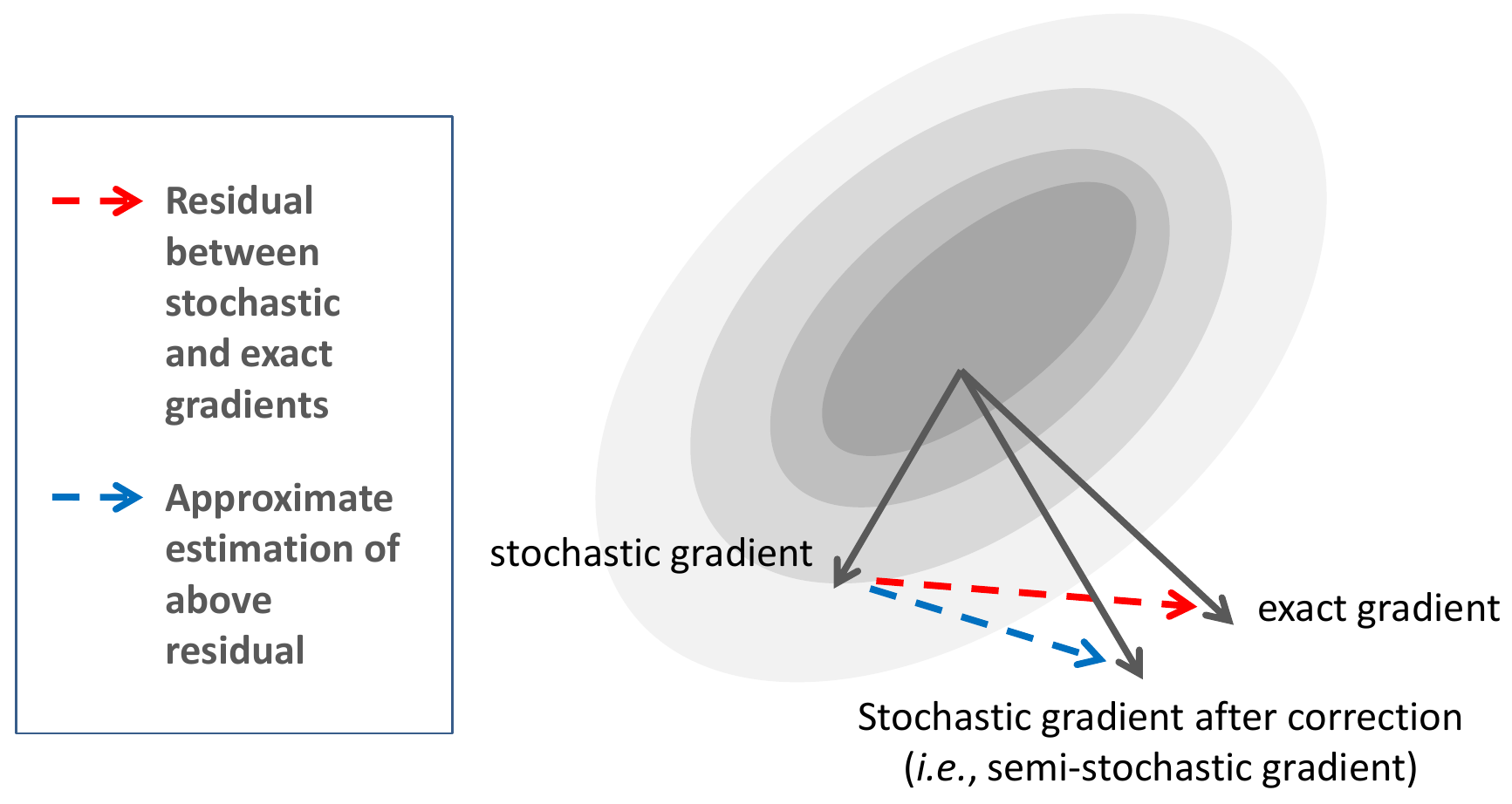}
\end{center}
   \caption{\small Illustration of residual-minimizing gradient correction. Stochastic gradient calculated from a single random sample often significantly deviates from the exact gradient. A simple solution is to compensate the stochastic gradient with the residual between the noisy stochastic gradient and full gradient (plotted as red dotted arrow in this figure). Exact residual is computationally expensive. Instead, semi-stochastic gradient descent approximately estimates the residue (plotted as blue dotted arrow in this figure), and amends the stochastic gradient accordingly. Best viewing in color mode.}
   \vspace{-0.1in}
\label{fig:idea}
\end{figure}

This paper advocates an efficient manifold propagation approach for reducing the stochastic gradient variance in large-scale machine learning. It aims to improve the stability of the stochastic gradient, such that large descending step sizes can be used for faster convergence. We adopt the computational framework of \emph{residual-minimizing gradient correction} which was originally proposed in \emph{stochastic variance-reduced gradient} (SVRG)~\cite{Johnson013} by Johnson and Zhang. The computational framework is comprised of two steps: 1) estimate the residual between a stochastic gradient and the full gradient using global information, and 2) compensate the stochastic gradient such that the residual is largely minimized.

Since the optimization proceeds in rounds, we can thus describe it with an update rule. For simplicity, consider the case that each mini-batch only contains a single random sample. Assume $\w^k$ is the latest estimation for the problem $\min_{\w}~ F(\w)$ at the $k$-th iteration, standard SGD and full (sub)gradient descend (GD) will seek for a new estimation $\w^{k+1}$ according to\footnote{\small When $F(\w)$ is non-smooth, sub-differential (rather than gradient) will be used. However, we here abuse the notation $\nabla$ for statement conciseness.}
\begin{eqnarray}
\textbf{(SGD)}: &&\w^{k+1} = \w^{k} - \eta_k  \nabla F_i(\w^{k}), \label{eqn:sgd} \\
\textbf{(GD)}: && \w^{k+1} = \w^{k} - \eta_k  \nabla F(\w^{k}), \label{eqn:gd}
\end{eqnarray}
where $\eta_k$ is a delicately-chosen step size. The term $F_i(\w^{k})$ in SGD denotes the atomic function conditioned on a random sample $\x_i$ and the latest parameters $\w^{k}$. $F(\w^{k})$ is computed using all training set.

In contrast, semi-stochastic gradient is obtained by the rule below:
\begin{eqnarray}
\w^{k+1} = \w^{k} - \eta_k \underbrace{\left( \nabla F_i(\w^{k}) -\left( \nabla F_i(\widetilde \w) - \nabla F(\widetilde \w) \right) \right)}_{semi-stochastic~gradient},\label{eqn:svrg}
\end{eqnarray}
where $\widetilde \w$ represents some historic memory of recent parameter estimation. $\widetilde \w$ is supposed to be proximal to $\w^{k}$. The term $\nabla F_i(\widetilde \w) - \nabla F(\widetilde \w)$ approximately estimates the residual between the stochastic gradient of sample $\x_i$ and full gradient. By subtracting the residual term from $\nabla F_i(\w^{k})$, it naturally aligns the stochastic gradient with the full gradient. As an extreme case, letting $\widetilde \w = \w^k$ will immediately get the full gradient in~(\ref{eqn:svrg}). The idea is intuitively explained in Figure~\ref{fig:idea}.

Theoretic analysis in~\cite{Johnson013,KonecnyR13,Xiao014} reveals that semi-stochastic algorithms achieve a geometric rate of convergence. Though such a convergence rate is generally regarded as the synonym of satisfactory efficiency, it is important to emphasize that this rate is achieved at the cost of higher iteration complexity compared to standard SGD. In our experiments, we are surprised to find that SGD still dominates in many cases, since its light-weight iteration cost compensates its slow theoretic convergence rate. In other words, the promising geometric convergence rate of existing semi-stochastic algorithms is probably Pyrrhic victories at excessive costs of maintaining high-accuracy estimation of gradient residual.

We find that a comprehensive quantitative comparison between semi-stochastic algorithms and SGD is still missing in the literature. In fact, most existing semi-stochastic algorithms either rely on periodic full gradient computation~\cite{Johnson013} or use Hessian-like covariance matrix operations~\cite{WangCSX13}, which account for their high iteration complexities. In this paper we expose a novel way of efficiently computing semi-stochastic gradient and evaluate it on a variety of massive data sets. We term the new method as \emph{stratified semi-stochastic gradient descent} (S3GD) hereafter. Our major contributions are described below:
\begin{itemize}
\item As a crucial component of the proposed S3GD, we devise an efficient manifold propagation approach for computing semi-stochastic gradient. First, a fixed number of anchors are drawn in a stratified manner. After that, each sample in the training set is connected to its adjacent anchors, forming a graph-defined manifold. At each iteration, the gradient information computed on the anchors diffuses over the manifold, obtaining an approximate estimation of the full gradient. The idea empirically proves to be a strong competitor to the existing expensive, albeit accurate, gradient-correcting operations such as SVRG.

\item We provide theoretic analysis about S3GD. Under standard assumptions imposed on the objective functions (\textit{i.e.}, strong convexity and Lipschitz continuity) and with a constant step size, the objective value obtained by S3GD converges to $F(\w^\ast) + \Delta$ at a geometric rate, where $F(\w^\ast)$ is the global minimum of $F(\w)$ and $\Delta$ is some quantity determined by the quality of anchor-based function approximation.

\item Last but not least, we conduct quantitative investigation over 9 different benchmarks, covering a large spectrum of real-world problems. The experimental evaluations fully validate the efficiency and effectiveness that S3GD brings. Moreover, the comparisons between various semi-stochastic algorithms and classic SGD is so far the most comprehensive and supposed to be very useful for re-calibrate the research direction of semi-stochastic algorithms.

\end{itemize}

The remainder of this paper is organized as follows: We start in Section~\ref{sec:algo} by describing preliminary knowledge and algorithmic details of S3GD. Specifically, Section~\ref{sec:instance} is devoted to applying the generic idea of S3GD to several representative machine learning problems. We then give the theoretic analysis in Section~\ref{sec:analysis}, where the major observation is found in Theorem~\ref{theorem:main}. In Section~\ref{sec:exp} we present the quantitative investigation of S3GD on several large-scale benchmark datasets widely used in machine learning and statistics. Finally, in Section~\ref{sec:conclusion} we draw the concluding remarks and discuss the future perspective.

\section{The Proposed Algorithm}
\label{sec:algo}

\subsection{Notations and Assumptions}


\textbf{Notations}: We will denote vectors and matrices by bold-face letters. Let $\| \x \|_2, \| \x \|_1$ be the Euclidean norm and 1-norm (summation of all absolute elements) of a vector respectively. $(x)_+ = \max(x,0)$ is the zero-thresholding operation. Denote the training data set as $\mathcal{X} = \{(\x_i,y_i)\}$, where $i = 1,\ldots,n$. Each sample is described by a tuple $(\x_i,y_i)$, where $\x_i \in \mathbb{R}^d$ is the feature vector and $y_i$ corresponds to either labels in supervised learning or response values in regression problems. The smooth part in Problem~(\ref{eqn:com}) is premised in an additive form, namely $P(\w) = (1/n) \sum_{i=1}^n \psi(\x_i, y_i, \w)$\footnote{\small $P(\w^\top \x)$ and $P(\w)$ will be interchangeably used in this paper. $P(\w^\top \x)$ will be used when we highlight the interplay between $\w$ and $\x$. Likewise $\psi_i(\w)$ and $\psi(\w^\top \x_i)$ are also equivalently used.}. The regularization term $R(\w)$ is convex yet not mandatorily differentiable. Whenever not incurring confusion, we use the notation $\psi_i(\w)$ for simplifying $\psi(\x_i, y_i, \w)$. Throughout this paper, by default we use $\| \x \|$ to represent the Euclidean norm unless otherwise clarified.

Our theoretic observations are based on the following assumptions, similar to previous semi-stochastic gradient descent methods~\cite{Xiao014,ZhaoZ14b}:
\begin{assumption}
\textbf{(strong convexity)}: \ \ We say that a function $f~:~\mathbb{R}^d \mapsto \mathbb{R}$ is strongly convex, if there exists $\mu>0$ such that for all $\u,\v \in \mathbb{R}^d$,
\begin{eqnarray}
\textstyle f(\u) \ge f(\v) + \xi^\top (\u-\v) + \frac{\mu}{2} \| \u - \v \|^2,~\forall \xi \in \partial f(\v),
\end{eqnarray}
where $\partial f(\v)$ is the sub-differential (set of sub-gradients) at point $\v$. The convexity parameter is defined to be the largest $\mu$ that satisfies the above condition. Let $P(\w),R(\w)$ and their composition $F(\w)$ have non-negative convexity parameters $\mu_P$, $\mu_R$ and $\mu$ respectively. It is easily verified that $\mu \ge \mu_p + \mu_R$ by definition of strong convexity and function composition.
\label{assumption:1}
\end{assumption}

\begin{assumption}
\textbf{(smoothness)}: A function $f~:~\mathbb{R}^d \mapsto \mathbb{R}$ is $L$-smooth if it is differential and there exists a smallest $L>0$ such that it satisfies
\begin{eqnarray}
\textstyle f(\u) \le f(\v) + \nabla f(\u) (\u-\v) + \frac{L}{2} \| \u - \v \|^2,
\end{eqnarray}
for all $\u,\v \in \mathbb{R}^d$. Or equivalently, its gradient is $L$-Lipschitz continuous, namely we have
\begin{eqnarray}
\| \nabla f(\u) - \nabla f(\v) \| \le L \| \u - \v \|.
\end{eqnarray}
Let the Lipschitz parameter for each atomic function $\psi_i(\w)$ be $L_i$ respectively. The Lipschitz parameter for their composition $P(\w)$ is $L_P \le (1/n) \sum_{i=1}^n L_i$. The regularization term $R(\w)$ is mostly assumed to be non-differentiable and thus has no Lipschitz parameter.
\label{assumption:2}
\end{assumption}

\subsection{Algorithmic Framework}

\begin{algorithm}[tb]
   \caption{The S3GD Algorithm}
   \label{alg:s3gd}
   \begin{small}
\begin{algorithmic}[1]
   \STATE {\bfseries Parameters:} maximal number of inner iterations $k_{in}$, the number of samples in a mini-batch $p$ and the step-size parameter $\eta$;
   \STATE {\bfseries Output:} optimal parameter vector $\w^\ast$;
   \STATE Initialize $\widetilde \w = \mathbf{0}$;
   \WHILE{not converged}
   \STATE $\w^0 =\widetilde \w$;
   \STATE Calculate the approximate full gradient $\nabla H(\widetilde \w)$ over the manifold according to Eqn.~(\ref{eqn:HM});
   \FOR{$k=1$ {\bfseries to} $k_{in}$}
   \STATE Construct a mini-batch by random sampling. Denote the index set as $\mathcal{I} = \{k_1,\ldots,k_p\}$.
   \STATE Calculate the stochastic gradient for the mini-batch, obtaining $\nabla P_\mathcal{I}(\w^{k-1}) = (1/p) \sum_{i=1}^p \nabla \psi_{k_i}(\w^{k-1})$;
   \STATE Calculate approximate stochastic gradient for the mini-batch on the manifold by Eqn.~(\ref{eqn:hi}), obtaining $\nabla h_\mathcal{I}(\widetilde \w) = (1/p) \sum_{i=1}^p \nabla h_{k_i}(\widetilde \w)$;
   \STATE Calculate the semi-stochastic gradient $g(\w^{k-1})$ according to Eqn.~(\ref{eqn:sgd_gb});
   \STATE Solve the following sub-problem:
   \begin{eqnarray}\w^{k} = \arg \min_{\w} ~\frac{1}{2} \| \w - (\w^{k-1} - \eta \cdot g(\w^{k-1})) \|_2^2 + \eta R(\w).\nonumber \end{eqnarray}
   \ENDFOR
   \STATE $\widetilde \w \leftarrow \w^{k_{in}}$;
   \ENDWHILE
   \STATE $\w^\ast \leftarrow \widetilde \w$;
\end{algorithmic}
\end{small}
\end{algorithm}

The composite optimization problem in~(\ref{eqn:com}) is of broad interests in machine learning and data mining fields. Nonetheless, solving it at optimal convergence speed is non-trivial. If we simply treat $F(\w)$ as a black-box oracle which only returns the first-order (sub)gradient, there are several off-the-shelf tools, including SGD and full (sub)gradient descent. Since full (sub)gradient estimation is extremely expensive when huge volume of data is available, recent work has focused on stochastic optimization.

SVRG~\cite{Johnson013}, as introduced in preceding section, obeys the update rule in~(\ref{eqn:svrg}). Procedurally, it utilizes two nested loops.
At each iteration of the outer loop, it memorizes a recent estimation $\widetilde \w$ and calculates the full gradient $\nabla F(\widetilde \w)$ at $\widetilde \w$. In the inner loops, it calculates $\nabla F_i(\w^{k})$ and $\nabla F_i(\widetilde \w)$ for mini-batches, and afterwards amends the stochastic gradient $\nabla F_i(\w^{k})$ by the rule in~(\ref{eqn:svrg}). Note that the same $\widetilde \w$ is used for all updates within an outer loop. The SVRG method, though simple, profoundly reduces the amortized time complexity at iterations and theoretically achieves geometric convergence rate for strongly-convex smooth functions.

Another semi-stochastic algorithm, \emph{stochastic control variate} (SCV)~\cite{WangCSX13},  represents a general approach
of using \emph{control variate} for reducing the variance of stochastic gradients. The update rule of SCV is similar to~(\ref{eqn:svrg}) yet the last two (sub)gradients in~~(\ref{eqn:svrg}) are replaced by control variate. Data statistics
such as low-order moments (vector mean and covariance matrix) can be used to form
the control variate. The authors apply SCV to solve logistic regression and latent Dirichlet allocation.

However, existing semi-stochastic methods like SVRG and SCV are not guaranteed to beat standard SGD in practice, since computing $\nabla F(\widetilde \w)$ in SVRG or control variate in SCV significantly increases the iteration complexity. To overcome the key limitations that dramatically restrict their capability in large scale data analysis, we propose S3GD. Algorithm~\ref{alg:s3gd} sketches the pseudo-code of S3GD.

Before diving into algorithmic details, we want to highlight two defining traits of S3GD:

\vspace{0.07in}
\noindent \textbf{Manifold-oriented gradient approximation:} Given the composite function $F(\w)$, S3GD only computes the gradient on the smooth part $P(\w)$. For accelerating the computation of semi-stochastic gradient in~(\ref{eqn:svrg}), we argue that the key is to find a function $H(\w)$, whose design principals are:

\vspace{0.05in}
1) $\nabla H(\w)$ is a good surrogate to the full gradient of the smooth component $\nabla P(\w)$, namely $\nabla H(\w) \approx \nabla P(\w)$;

2) $\nabla H(\w)$ can be efficiently computed;

3) $\nabla H(\w) = \frac{1}{n} \sum_{i=1}^n \nabla h_i(\w)$ is additive, where $\nabla h_i(\w)$ approximates the stochastic gradient of an atomic function. Namely, $\nabla h_i(\w) \approx \nabla \psi_i(\w)$.
\vspace{0.05in}

We defer the construction of function $H(\w)$ in Section~\ref{subsec:sga}, focusing on the algorithmic pipeline here. At specific iteration, an index set $\mathcal{I} \subset \{1,\ldots,n\}$ is randomly generated for constructing a mini-batch. Conditioned on current parameter estimation ${\w}^{k}$ and a recent historical estimation $\widetilde \w$, the semi-stochastic gradient in S3GD is computed by the following formula:
\begin{eqnarray}
g_\mathcal{I}(\w^{k},\widetilde \w) = \nabla \psi_\mathcal{I}({\w}^{k}) - \left[\nabla h_\mathcal{I}(\widetilde \w) - \nabla H(\widetilde \w) \right],
\label{eqn:sgd_gb}
\end{eqnarray}
where $\nabla \psi_\mathcal{I}({\w}^{k}) = \sum_{i \in \mathcal{I}} \nabla \psi_i(\w^k) / |\mathcal{I}|$, $\nabla h_\mathcal{I}(\widetilde \w) = \sum_{i \in \mathcal{I}} \nabla h_i(\widetilde \w) / |\mathcal{I}|$ are the averaged original / approximate stochastic gradients over the index set $\mathcal{I}$ respectively. Hereafter we use $g_\mathcal{I}(\w^{k})$ for brevity since the parameter vector $\widetilde \w$ can be mostly inferred from the context.

In fact, $g_\mathcal{I}(\w^{k})$ provides an unbiased estimate of $\nabla P(\w^{k})$ when $\mathcal{I}$ is randomly drawn from $[1,\ldots,n]$ without replacement. Its soundness is naturally fulfilled by the additive construction of functions $P(\w)$ and $H(\w)$. Consequently, the variance of $g_\mathcal{I}(\w^{k})$ becomes
\begin{eqnarray}
Var\left[ g_\mathcal{I}(\w^{k}) \right] &=& \mathbb{E} \left \| \nabla \psi_\mathcal{I}({\w^{k}}) - \nabla h_\mathcal{I}(\widetilde \w) \right \|^2 \nonumber \\
&&- \left( \mathbb{E} \left \| \nabla P(\w^{k}) - \nabla H(\widetilde \w) \right \| \right)^2 \nonumber \\
&\le& \mathbb{E} \left \| \nabla \psi_\mathcal{I}({\w^{k}}) - \nabla h_\mathcal{I}(\widetilde \w) \right \|^2. \label{eqn:var1}
\end{eqnarray}

In comparison, the variance of noisy stochastic gradient $\nabla \psi_\mathcal{I}({\w^{k}})$ in standard SGD is
\begin{eqnarray}
Var\left[ \nabla \psi_\mathcal{I}({\w^{k}}) \right] = \mathbb{E} \left \| \nabla \psi_\mathcal{I}({\w^{k}}) - \nabla P({\w ^{k}}) \right \|^2. \label{eqn:var2}
\end{eqnarray}

For $Var[ g_\mathcal{I}(\w^{k}) ]$ and $Var[ \nabla \psi_\mathcal{I}({\w^{k}}) ]$, the smaller one is more favorable. As shown later, to reduce $Var\left[ g_\mathcal{I}(\w^{k}) \right]$, we designate $\nabla h_\mathcal{I}(\widetilde \w)$ to be a localized approximation of $\nabla \psi_\mathcal{I}(\w^{k})$. It is supposedly closer to $\nabla \psi_\mathcal{I}({\w^{k}})$ in comparison with the global average $\nabla P({\w^{k}})$, particularly when the input data set is with rich variety.

\vspace{0.07in}
\noindent \textbf{Proximity-regularized linear approximation:} After the semi-stochastic gradient $g_\mathcal{I}(\w^{k})$ is computed, we further solve the following sub-problem:
\begin{eqnarray}
\w^{k+1} = \arg \min_{\w} && \textstyle P(\w^{k}) + \langle g_\mathcal{I}(\w^{k}), \w -  \w^{k} \rangle \nonumber \\
&& \textstyle \frac{1}{2 \eta} \left \| \w - \w^{k} \right \|^2 + R(\w),
\label{eqn:wk}
\end{eqnarray}
where the first three terms define a proximal regularization of the linearized approximation of $P(\w)$ around point $\w^{k}$. $R(\w)$ is presumably in a good shape such that solving (\ref{eqn:wk}) is trivial. If $R(\w)$ is itself composition of several non-smooth functions, one can resort to the modern proximal average techniques~\cite{Yu13a}. Moreover, it is verified that Problem~(\ref{eqn:wk}) can be compactly abstracted by the operation $\mathbf{prox}()$ below:
\begin{eqnarray}
\textstyle \mathbf{prox}_{\eta R}(\u) = \arg \min_{\w} ~ \frac{1}{2} \| \w - \u \|^2 + \eta R(\w), \label{eqn:wk2}
\end{eqnarray}
where $\u = \w^{k} - \eta \cdot g_\mathcal{I}(\w^{k})$.

\subsection{Gradient Approximation by Manifold Propagation}
\label{subsec:sga}

\begin{algorithm}[tb]
   \caption{Manifold Based Gradient Approximation}
   \label{alg:anchor}
   \begin{small}
\begin{algorithmic}[1]
   \STATE {\bfseries Parameters:} anchor number $m$ and $k$-NN parameter $k$.
   \newline 
   {\bfseries \underline{Anchor Selection}}
   \vspace{0.04in}
   \STATE Perform data clustering to obtain $m$ centers $\c_i$, $i=1\ldots m$;
   \FOR{$i=1$ {\bfseries to} $m$}
   \STATE Find anchor $\z_i$ by solving
   \begin{eqnarray}
   \z_i = \arg \min\nolimits_{\x} \quad \| \x - \c_i \|^2,
   \end{eqnarray}
   where $\x$ is from the training data set.
   \ENDFOR
   \newline 
   {\bfseries \underline{Sparse Anchor-Sample Graph (ASG) Construction}}
   \vspace{0.04in}
   \FOR{$i=1$ {\bfseries to} $n$}
   \STATE For sample $\x_i$, find $k$-nearest anchors $\z_{i_1},\ldots,\z_{i_k}$;
   \STATE Learn the Gaussian kernel parameter by
   \begin{eqnarray}
   \sigma = \max \left(\epsilon, \inf_{j \in \{i_1,\ldots,i_k\}} \sqrt{\|x_i - \z_j\|} \right),
   \end{eqnarray}
   where $\epsilon$ is set to be $10^{-4}$ to avoid the trivial case $\sigma=0$.
   \FOR{each $k$-nearest anchor $\z$}
   \STATE Calculate $\gamma_{\z}(\x_i) = \exp(-\|\x_i-\z\|^2/\sigma^2)$;
   \ENDFOR
   \STATE Normalize $\gamma_{\z}(\x_i)$ to ensure that they sum to 1;
   \ENDFOR
   \newline 
   {\bfseries \underline{Gradient Approximation over ASG}}
   \vspace{0.04in}
   \STATE Pre-compute the product matrix $\X \M$ in Eqn.~(\ref{eqn:HM});
   \vspace{0.08in}
   \STATE // for $\nabla h_\mathcal{I}(\w)$ on a mini-batch
   \FOR{each $\x_i$ in the mini-batch $\mathcal{I}$}
   \STATE Calculate $\nabla h_i(\w)$ by Eqn.~(\ref{eqn:hi});
   \ENDFOR
   \STATE $\nabla h_\mathcal{I}(\w) = \sum_{i \in \mathcal{I}} \nabla h_i(\w) / |\mathcal{I}|$;
   \vspace{0.08in}
   \STATE // for approximate full gradient $\nabla H(\w)$
   \STATE Calculate $\nabla H(\w)$ by Eqn.~(\ref{eqn:HM});
\end{algorithmic}
\end{small}
\end{algorithm}

This section elaborates on a manifold-oriented method for approximating the stochastic gradient $\nabla \psi_i(\w)$. Our key argument is that a universal gradient-approximating function is either infeasible or inaccurate in general. Our proposed solution is anchor-based gradient approximation over non-linear data manifold. The idea has ever been explored in other context (such as feature dimension reduction) yet not in stochastic optimization before.

Yu et al. showed in~\cite{YuZG09} that any Lipschitz-continuous function $f(\x)$ residing on lower-dimensional manifolds can be approximated by a linear combination
of function values, namely
\begin{eqnarray}
f(\x) \approx \sum_{\z \in \mathcal{Z}} \gamma_{\z} (\x) f(\z), \label{eqn:fx}
\end{eqnarray}
where $\mathcal{Z}$ is a collection of pre-specified anchors. $\gamma_{\z} (\x) \ge 0$ is the combination coefficient depending on both the data vector $\x$ and anchor $\z$. The idea is later generalized in the work of \emph{locally-linear support vector machine}~\cite{LadickyT11,FornoniCO13}, where each anchor determines a function (rather than a fixed value), namely $f(\z)$ in~(\ref{eqn:fx}) is replaced by an $\x$-varying function $f_{\z}(\x)$.

In Problem~(\ref{eqn:com}), we assume that each atomic loss function $\psi(\w^\top \x_i)$ is linear with respect to $\x_i$. Letting $\psi^\prime(u)$ be the derivative with respect to a scalar $u$, the stochastic gradient of $\x_i$ with respect to $\w$ can be factorized as below:
\begin{eqnarray}
\nabla \psi_i(\w) = \psi^\prime(\w^\top \x_i) \cdot \x_i. \label{eqn:np}
\end{eqnarray}

Inspired by the factorization in~(\ref{eqn:np}), we propose to establish a manifold over the training data, such that the derivative term $\psi^\prime(\w^\top \x_i)$ in~(\ref{eqn:np}) can be efficiently calculated via sparse information propagation on the manifold. Algorithm~\ref{alg:anchor} shows the pseudo-code for the major steps. The proposed scheme consists of the following components:

\vspace{0.05in}
\textbf{1) Constructing anchor set}: Compared to universal gradient approximation, anchor set~\cite{LiuHC10} has a stronger representation power by establishing local approximation around each anchor point. Let $m$ be the number of anchor points, whose optimal value of is mostly dataset-specific. Let $\mathcal{Z} = \{\z_1,\ldots,\z_m\}$ be the anchor set. We employ a $k$-means clustering procedure to obtain $m$ centers in a stratified manner. The anchor points are chosen as the nearest samples to these centers, since these centers per se are not necessarily corresponding to meaningful features.

\vspace{0.05in}
\textbf{2) Anchor-Sample Graph (ASG) Construction}: We follow the local approximation scheme as described in Eqn.~(\ref{eqn:fx}). Specifically, we propose to approximate the term $\psi^\prime(\w^\top \x_i)$ in~(\ref{eqn:np}) by
\begin{eqnarray}
\psi^\prime(\w^\top \x_i) \approx \sum_{\z \in \mathcal{Z}} \gamma_{\z}(\x_i) \cdot \psi^\prime(\w^\top \z). \label{eqn:psip}
\end{eqnarray}

Each anchor $\z$ uniquely determines a localized function $\psi^\prime(\w^\top \z)$, whose value varies with respect to different $\w$. The coefficient $\gamma_{\z}(\x_i)$ controls the contribution of specific anchor point in computing $\psi^\prime(\w^\top \x_i)$. Geometrically, anchors and all training samples naturally form an \emph{anchor-sample graph} (ASG), where the connectivity strengths are controlled by $\{\gamma_{\z}(\x)\}$. In graph-based propagation methods, it is known that connecting sample with remote anchors potentially does harm to the performance~\cite{LiuHC10}. Therefore, each sample is enforced to only connect to its $k$-nearest anchors. State differently, most $\gamma_{\z}(\x)$ is zero so that the ASG is highly sparse. The computation of $\gamma_{\z}(\x)$ is detailed in Algorithm~\ref{alg:anchor}.

\vspace{0.05in}
\textbf{3) Gradient Approximation over ASG}: Combining Eqns.~(\ref{eqn:np}) and~(\ref{eqn:psip}) obtains
\begin{eqnarray}
\nabla \psi_i(\w) &=& \psi^\prime(\w^\top \x_i) \cdot \x_i \nonumber \\
&\approx& \textstyle \left( \sum_{\z \in \mathcal{Z}} \gamma_{\z} (\x_i) \cdot \psi^\prime(\w^\top \z) \right) \cdot \x_i,
\label{eqn:appgrd}
\end{eqnarray}
where the right hand side in~(\ref{eqn:appgrd}) serves as our proposed manifold-oriented approximate gradient. Formally, we designate a surrogate function $h_i(\w)$ such that
\begin{eqnarray}
\nabla h_i(\w) = \left( \sum_{\z \in \mathcal{Z}} \gamma_{\z} (\x_i) \cdot \psi^\prime(\w^\top \z) \right) \x_i. \label{eqn:hi}
\end{eqnarray}

Likewise, the approximate full gradient $\nabla H(\w)$ in~(\ref{eqn:sgd_gb}) can be computed by:
\begin{eqnarray}
\nabla H(\w) &=&  \frac{1}{n} \sum_{i=1\ldots n} \nabla h_i(\w) \nonumber \\
&=&  \frac{1}{n} \sum_{i=1\ldots n} \left[ \left( \sum_{\z \in \mathcal{Z}} \gamma_{\z} (\x_i) \cdot \psi^\prime(\w^\top \z) \right) \cdot \x_i \right] \nonumber \\
&=&  \frac{1}{n} \sum_{\z \in \mathcal{Z}} \left[ \sum_{i=1\ldots n} \gamma_{\z} (\x_i) \cdot \x_i \right] \cdot \psi^\prime(\w^\top \z). \label{eqn:H}
\end{eqnarray}

Importantly, computing~(\ref{eqn:hi}) and~(\ref{eqn:H}) is highly efficient owing to the sparsity of ASG. It only involves executing the derivative function for all anchors in addition to another $\mathcal{O}(m+d)$ algorithmic operations per sample. In fact, the computation in~(\ref{eqn:H}) can be further accelerated by per-computing the terms irrelevant to $\w$. Let $\M \in \mathbb{R}^{n \times m}$ be the matrix by compiling all coefficients in ASG. Specifically, $\M(i, j) = \gamma_{\z_j}(\x_i)$. Moreover, let
\begin{eqnarray}
\d_{\mathcal{Z}}(\w) = \left( \psi^\prime(\w^\top \z_1),\ldots, \psi^\prime(\w^\top \z_m) \right)^\top \in \mathbb{R}^m, \label{eqn:21}
\end{eqnarray}
be the vector of anchor derivatives conditioned on parameter $\w$. $\X = (\x_1,\ldots,\x_n) \in \mathbb{R}^{d \times n}$ is the feature matrix. Eqn.~(\ref{eqn:H}) can be compactly expressed as
\begin{eqnarray}
\nabla H(\w) = \frac{1}{n}  \cdot (\X \M) \times \d_{\mathcal{Z}}(\w), \label{eqn:HM}
\end{eqnarray}

The product $\X \M \in \mathbb{R}^{d \times m}$ is not varying with respect to $\w$ and thus can be pre-computed for avoiding redundant computation at different outer loops in Algorithm~\ref{alg:s3gd}.

\subsection{Instances of Applications}
\label{sec:instance}

This section instantiates our proposed algorithm by several representative loss functions and regularizations.

\vspace{0.05in}
\noindent \textbf{Logistic Loss}: It is applicable to either real or binary responses. We focus on the binary case, where the label $y=\pm 1$. For any data vector $\x$, the conditional probability of the class label is:\footnote{\small For the sake of simplifying notations, we remove the intercept variable by appending an additional dimension of constant 1 to any feature vector $\x$.}
\begin{eqnarray}
p(y | \x; \w) = \sigma(y \w^\top \x) := 1 / (1 + \exp(-y (\w^\top \x) ) ).
\end{eqnarray}

The log-likelihood function is then expressed as $P(\w) = \sum_{i=1}^n \psi(\w^\top \x_i) = \sum_{i=1}^n \log p(y_i | \x_i; \w)$. According to the calculus rule of sigmoid function, the gradient of $\psi(\w^\top \x_i)$ is:
\begin{eqnarray}
\nabla \psi(\w^\top \x_i) = \underbrace{\sigma(-y_i \w^\top \x_i)}_{\small{\mbox{derivative~}}\psi^\prime(\w^\top \x_i)} \cdot y_i \x_i \label{eqn:hinge}
\end{eqnarray}

Applying the idea of anchor-based approximation in Eqn.~(\ref{eqn:hi}), we have
\begin{eqnarray}
\nabla \psi(\w^\top \x_i) &=&  \sigma(-y_i \w^\top \x_i)  \cdot y_i \x_i  \nonumber \\
&\approx& \left( \sum_{\z \in \mathcal{Z}} \gamma_{\z} (\x_i) \cdot \sigma(-y_i \w^\top \z) \right) \cdot y_i \x_i, \nonumber
\end{eqnarray}
which indicates that the approximate stochastic gradient is
\begin{eqnarray}
\nabla h_i(\w) = \left( \sum_{\z \in \mathcal{Z}} \gamma_{\z} (\x_i) \cdot \sigma(- y_i \w^\top \z) \right) \cdot y_i \x_i. \label{eqn:ins_1}
\end{eqnarray}


The label $\y_i$ and feature vector $\z$ are tightly coupled in Eqn.~(\ref{eqn:ins_1}), which prohibits the matrix multiplication in Eqn.~(\ref{eqn:HM}). To decouple them, the stochastic gradients of different samples can be handled according to their labels. More formally, let us consider the following two cases:

\vspace{0.04in}
\emph{Case-1}: $y_i=1$. We have $\nabla \psi(\w^\top \x_i) = \sigma(- \w^\top \x_i) \cdot \x_i$, where $\sigma(- \w^\top \x_i)$ can be efficiently approximated by $\sum_{\z} \gamma_{\z} (\x_i) \cdot \sigma(- \w^\top \z)$.

\vspace{0.04in}
\emph{Case-2}: $y_i=-1$. Now there is $\nabla \psi(\w^\top \x_i) = \sigma( \w^\top \x_i) \cdot (-\x_i) = (1-\sigma( -\w^\top \x_i)) \cdot (-\x_i) = -\x_i + \sigma(- \w^\top \x_i) \cdot \x_i$. Note that we use the property of sigmoid function $\sigma(u) = 1 - \sigma(-u)$. It turns out that we can still apply the tricks in Case-1 by amending the result with an additional term $-\x_i$. In practice, to estimate $\nabla H(\w)$, we can pre-compute $(1/n) \sum_{i ~:~ y_i=-1} \x_i$ and use it to compensate the quantity calculated from Eqn.~(\ref{eqn:HM}). This way the computation still enjoys the high efficacy of matrix-based operations.

\vspace{0.05in}
\noindent \textbf{Hinge Loss and Squared Hinge Loss}: The loss function popularized by SVM is known to be hinge loss $(1-y \w^\top \x)_+$. It is non-differentiable due to the irregularity at $y \w^\top \x = 1$. However, as discovered in~\cite{Zhangt01,ZhangJYH03}, hinge loss can be smoothed by the loss of ``modified logistic regression":
\begin{eqnarray}
(1- y \w^\top \x)_+ \approx \frac{1}{\beta} \log \left(1 + \exp(-\beta(y \w^\top \x -1)) \right).
\end{eqnarray}

The approximation residual asymptotically becomes zero when $\beta \rightarrow +\infty$, therefore we can cast hinge loss into the the framework of logistic loss with properly-chosen $\beta$.

Another solution of smoothing hinge loss is using squared hinge loss as adopted by L2-SVM~\cite{Chapelle}, namely $(1/2)((1-y \w^\top \x)_+)^2$, which naturally removes the irregular point at the risk of over-penalizing large response. Its gradient at a sample $(\x_i,y_i)$ is
\begin{eqnarray}
\nabla \psi(\w^\top \x_i) = \underbrace{ (1- y_i \w^\top  \x_i)_+  }_{\small{\mbox{derivative~}}\psi^\prime(\w^\top \x_i)} \cdot (-y_i \x_i).
\end{eqnarray}


\vspace{0.05in}
\noindent \textbf{Regularization}: The regularization function $R(\w)$ can be either smooth (\emph{e.g.}, Tikhonov regularization) or non-smooth (\emph{e.g.}, 1-norm regularization). Below we list a few regularization functions widely-used in machine learning:
\begin{eqnarray}
(\mbox{Tikhonov}) : &&R(\w) = \lambda \| \w \|^2_2. \nonumber \\
(\mbox{1-norm}): &&R(\w) = \lambda  \| \w \|_1. \nonumber \\
(\mbox{Elastic net}): &&R(\w) = \lambda (1-\alpha) \| \w \|_1 + \lambda \alpha \| \w \|_2. \nonumber
\end{eqnarray}

When parameters $\w$ constitute a matrix rather than a vector, regularization terms such as matrix nuclear norm~\cite{svt} can be applied. However, optimizing with all above regularization under the proximal operator in~(\ref{eqn:wk2}) has been maturely developed. We thus omit more discussion.

\subsection{Algorithmic Complexity}

\begin{table*}[t]
\centering
\begin{tabular}{crrrrr}
  \hline\hline
 \textbf{} & \textbf{SGD} & \textbf{SSGD} & \textbf{SVRG}   & \textbf{SCV} & \textbf{S3GD} \\
  \hline
\textbf{Iteration Time Complexity:} & $\mathcal{O}(p \times d)$  & $\mathcal{O}(p \times d)$ & $\mathcal{O}\left(p \times d + \frac{n}{k_{in}} \times d \right)$   & $\mathcal{O}\left(p \times d + d^2 \right)$ & $\mathcal{O}\left(p \times d + \frac{m}{k_{in}} \times d + p \times (k+d) \right)$\\
\hline
\textbf{Preprocessing Time Complexity:} & --  & $\mathcal{O}(n \times m \times d)$ & --  & -- & $\mathcal{O}(n \times m \times d)$\\
\hline
\textbf{Space Complexity:} & $\mathcal{O}(d)$  & $\mathcal{O}(d m) + \mathcal{O}(n)$ & $\mathcal{O}(d)$  & $\mathcal{O}(d^2)$ & $\mathcal{O}(d m) + \mathcal{O}(n k)$\\
\hline

\end{tabular}
\caption{\small Time/space iteration complexity for all algorithms involved in the quantitative study. $p, m, n, d$ denote the size of a mini-batch, the number of anchors in S3GD (or the number of clusters in SSGD), the number of training samples, and the feature dimensionality respectively. $k$ denotes the anchor $k$-NN parameter. Note that for SVRG and S3GD, they both adopt nested loop during the optimization. $k_{in}$ denotes the maximal iteration count of the inner loop. The mark ``--" implies the absence of any pre-processing.}
\label{table:complexity}
\end{table*}

The iteration complexity of the proposed S3GD depends on several parameters: the mini-batch size $p$, the number of anchors $m$, the $k$-NN parameter in constructing ASG, the maximal inner loop count $k_{in}$ and the feature dimensionality $d$. In the pre-computation, obtaining the matrix product $\X \M$ requires $\mathcal{O}(d k n)$ time complexity. And finding the $k$ nearest anchors has a time complexity of $\mathcal{O}(n m d)$. At the run-time, the complexity of computing $\nabla H(\widetilde \w)$ is comprised of $\mathcal{O}(d m)$ for evaluating anchor gradients in Eqn.~(\ref{eqn:21}) and $\mathcal{O}(d m)$ for multiplying the pre-computed $\X \M$ with a vector $d_{\mathcal{Z}}(\w)$. Note that $\nabla H(\widetilde \w)$ is only calculated once at the beginning of each outer loop in Algorithm~\ref{alg:s3gd}. Computing $\nabla \psi_\mathcal{I}({\w}^{k})$ and $\nabla h_\mathcal{I}(\widetilde \w)$ in Eqn.~(\ref{eqn:sgd_gb}) admits a time complexity of $\mathcal{O}(p d)$ or $\mathcal{O}(p (k + d))$ respectively.

Most of existing semi-stochastic algorithms rely on two nested loop, of which the outer loop incurs exact full gradient computation or covariance matrix estimation. For large data, it entails a tremendous $\mathcal{O}(n d)$ or $\mathcal{O}(d^2)$ complexity. For other sophisticated algorithms that target at improved mini-batch construction (such as SSGD~\cite{ZhaoZ14b}), the iteration complexity is generally better than ours. However, the lack of global information makes these algorithms more sensitive to noise in stochastic gradients.

Regarding the space requirement, the major costs for S3GD include $\mathcal{O}(d m)$ for storing the product matrix in Eqn.~(\ref{eqn:HM}) and $\mathcal{O} (n k)$ for recording $k$ nearest anchors for each sample. Akin to SVRG and SCV, S3GD does not memorize historic gradients. We summarize the space and time complexities for all interested algorithms in Table~\ref{table:complexity}.

\section{Convergence Analysis}
\label{sec:analysis}

We need two lemmas as below to advance the convergence analysis. The first lemma states
\begin{lemma}
\label{lemma:lemma1}
Consider the composite function $F(\w)$ as defined in Eqn.~(\ref{eqn:com}). It satisfies both Assumptions 2.1 and 2.2. Let $w^\ast = \arg \min_{\w} F(\w)$ be the optimal solution and $L = \max_i L_i$ be the maximal Lipschitz-continuous constant of $\psi_i(\w)$, $i=1,\ldots,n$. Then we have
\begin{eqnarray}
\frac{1}{n} \sum_{i=1}^n \left\| \nabla \psi_i(\w) - \nabla \psi_i(\w^\ast) \right\|^2 \le 2 L \left[ F(\w) - F(\w^\ast) \right].
\end{eqnarray}
\end{lemma}

The key tricks in the proof of Lemma~\ref{lemma:lemma1} were originally developed in~\cite{Johnson013} and a complete proof is found in Lemma 3.4 of~\cite{Xiao014}. The other lemma is an extension of above lemma with anchor-based gradient approximation.
\begin{lemma}
\label{lemma:lemma2}
Consider the same problem setting as in Lemma~\ref{lemma:lemma1}. The gradient of $\psi_i(\w)$ is approximated by $\nabla h_i(\w)$ according to Eqn.~(\ref{eqn:hi}). Then
\begin{eqnarray}
&& \frac{1}{n} \sum_{i=1}^n \left\| \nabla h_i(\w) - \nabla \psi_i(\w^\ast) \right\|^2 \nonumber \\
&\le& 2 L (F(\w) - F(\w^\ast))  + 4 L \max_{\w} \left| H(\w) - P(\w) \right|, \nonumber
\end{eqnarray}
where $H(\w) = \textstyle \frac{1}{n} \sum_{i=1}^n h_i(\w)$.
\begin{proof}
Let us construct an auxiliary function as below,
\begin{eqnarray}
a_i(\w) = h_i(\w) - \psi_i(\w^\ast) - \nabla \psi_i(\w^\ast)^\top (\w - \w^\ast),
\end{eqnarray}
where $i \in \{1,\ldots,n\}$. It is easily verified that $\nabla a_i(\w) = \nabla h_i(\w) - \nabla \psi_i(\w^\ast)$. From the construction of $h_i(\w)$, $a_i(\w)$ is Lipschitz continuous with constant $L = \max_i L_i$. Based on an argument in Theorem 2.1.5 in~\cite{Nesterov}, we have
\begin{eqnarray}
\frac{1}{2 L} \| \nabla a_i(\w) \|^2 \le a_i(\w) - \min_{\w} a_i(\w).
\end{eqnarray}

Averaging over $i = 1,\ldots,n$, we obtain
\begin{eqnarray}
&& \frac{1}{n} \sum_{i=1}^n \left\| \nabla h_i(\w) - \nabla \psi_i(\w^\ast) \right\|^2 \nonumber \\
&\le& 2 L \left[ \frac{1}{n} \sum_{i=1}^n a_i(\w) - \frac{1}{n} \sum_{i=1}^n \min_{\w} a_i(\w) \right]. \label{eqn:2_1}
\end{eqnarray}

From the definition of $a_i(\w)$, we have
\begin{eqnarray}
\frac{1}{n} \sum_{i=1}^n a_i(\w) = H(\w) - P(\w^\ast) - \nabla P(\w^\ast)^\top (\w - \w^\ast). \nonumber
\end{eqnarray}

By the optimality of $\w^\ast$, a sub-gradient $\xi^\ast \in \partial R(\w^\ast)$ exists such that $\nabla P(\w^\ast) + \xi^\ast = 0$. Therefore
\begin{eqnarray}
&& \frac{1}{n} \sum_{i=1}^n a_i(\w) \nonumber \\
&=& H(\w) - P(\w^\ast) + (\xi^\ast)^\top (\w - \w^\ast) \nonumber \\
&\le& H(\w) - P(\w^\ast) + R(\w) - R(\w^\ast) \nonumber \\
&=& H(\w) - P(\w) + F(\w) - F(\w^\ast) \nonumber \\
&\le& \max_{\w} \left| H(\w) - P(\w) \right| + F(\w) - F(\w^\ast) , \label{eqn:m4}
\end{eqnarray}
where the first inequality follows from the definition of sub-gradient.

The second term on the right hand side of~(\ref{eqn:2_1}) can be bounded as below:
\begin{eqnarray}
&& \frac{1}{n} \sum_{i=1}^n \min_{\w} a_i(\w) \nonumber \\
&\ge& \min_{\w}  \frac{1}{n} \sum_{i=1}^n a_i(\w)  \nonumber \\
&=& \min_{\w}  \frac{1}{n} \sum_{i=1}^n \left[ h_i(\w) - \psi_i(\w^\ast) - \nabla \psi_i(\w^\ast)^\top (\w - \w^\ast)  \right] \nonumber \\
&\ge& \min_{\w}  \frac{1}{n} \sum_{i=1}^n \left[ h_{i}(\w) - \psi_i(\w) \right]  \label{eqn:m5_a} \\
&=& \min_{\w} \left[ H(\w) - P(\w)\right] \nonumber \\
&\ge& - \max_{\w} \left| H(\w) - P(\w)\right|. \label{eqn:m5}
\end{eqnarray}

Combining~(\ref{eqn:m4}) and~(\ref{eqn:m5}) completes the proof.

\end{proof}
\end{lemma}

The following is our main observation regarding the convergence property of the proposed S3GD:
\begin{theorem} \label{thm:main}
For compositional function $F(\w) = P(\w) + R(\w)$, assume its two components $P(\w)$, $R(\w)$ have strong convexity parameter $\mu_P \ge 0, \mu_R \ge 0$ and $\mu_P \cdot \mu_R > 0$. $\psi_i(\w)$ has Lipschitz parameter $L_i$, $i=1,\ldots,n$. Let $L = \max_i L_i$ be the maximal Lipschitz-continuous constant of $\psi_i(\w)$. Set the step size $\eta  \in (0, \frac{1}{8 L})$ and the iteration count in the inner loop (denote it as $m$) sufficiently large, such that
\begin{eqnarray}
\rho = \frac{1}{(\mu_P+\mu_R) \eta (1-4 L \eta) m} + \frac{4 L \eta (m+1)}{(1 - 4 L \eta ) m} < 1,\label{eqn:main_1}
\end{eqnarray}
and
\begin{eqnarray}
\Delta = \frac{ 8  \eta^2 L (\mu_P + \mu_R) m  } {\eta (\mu_P+\mu_R) (m - 4 \eta L (2 m + 1)) - 1} \cdot \Delta_{H,P} > 0, \label{eqn:main_2}
\end{eqnarray}
where $\Delta_{H,P} = \max_{\w} | H(\w) - P(\w) |$ denotes the maximal residual between the smooth component $P(\w)$ and its anchor-based approximation $H(\w)$.

Moreover, use $s$ to index the outer loops. The proposed S3GD algorithm will satisfy the following inequality,
\begin{eqnarray}
F(\w_{s})  - F(\w^\ast) - \Delta \le \rho^s \left[ F(\w_0) - F(\w^\ast) -\Delta \right], \label{eqn:main_3}
\end{eqnarray}
where $\rho$ and $\Delta$ are defined in Eqns.~(\ref{eqn:main_1})(\ref{eqn:main_2}) respectively.
\label{theorem:main}
\begin{proof}
The proof is essentially an adaption of Theorem 3.1 in~\cite{Xiao014}. Let us consider a single outer loop (indexed by $s$) which consist of $m$ inner iterations in total. Use $\w^k$ to denote the parameter vector at the $k$-th inner iteration. $\w^0$ is initialized as $\w_s$ obtained in previous outer loop. Without loss of generality, let us consider mini-batches with single random sample at each inner iteration. Based on Lemma 3.7 in~\cite{Xiao014}, we have
\begin{eqnarray}
&& \| \w^{k}-\w^\ast \|^2 \nonumber \\
&\le& \| \w^{k-1}-\w^\ast \|^2 - \eta \mu_P \| \w^{k-1}-\w^\ast \|^2  \nonumber \\
&&  - \eta \mu_R \| \w^{k}-\w^\ast \|^2 -2 \eta ( F(\w^{k}) - F(\w^\ast) ) \nonumber \\
&& - 2 \eta \Delta_k^\top (\w^{k}-\w^\ast) \nonumber \\
&\le& \| \w^{k-1}-\w^\ast \|^2 -2 \eta ( F(\w^{k}) - F(\w^\ast) ) \nonumber \\
&& - 2 \eta \Delta_k^\top (\w^{k}-\w^\ast), \label{eqn:p1}
\end{eqnarray}
where we use the notations $\Delta_k = \v_k - \nabla P(\w^{k-1})$ and
\begin{eqnarray}
\v_k = \nabla \psi_{i_k}({\w}^{k-1}) - [\nabla h_{i_k}(\w_s) - \nabla H(\w_s)].\label{eqn:p344}
\end{eqnarray}

To further bound~(\ref{eqn:p1}), we have to investigate the term $\Delta_k^\top (\w^{k}-\w^\ast)$. To this end, let us define the proximal full gradient update as
\begin{eqnarray}
\widetilde \w^{k} = \mathbf{prox}_{\eta R}(\w^{k-1} - \eta \nabla P(\w^{k-1})).
\end{eqnarray}

An argument in Theorem 3.1 of~\cite{Xiao014} indicates that the relation below holds:
\begin{eqnarray}
-\Delta_k^\top (\w^{k}-\w^\ast) \le \eta \| \Delta_k \|^2 - \Delta_k^\top (\widetilde \w^{k} - \w^\ast). \label{eqn:p611}
\end{eqnarray}

Applying~(\ref{eqn:p611}) to~(\ref{eqn:p1}) obtains
\begin{eqnarray}
&& \| \w^{k}-\w^\ast \|^2 \nonumber \\
&\le& \| \w^{k-1}-\w^\ast \|^2 -2 \eta ( F(\w^{k}) - F(\w^\ast) )\nonumber \\
&&+ 2 \eta^2 \| \Delta_k \|^2 -  2 \eta \Delta_k^\top (\widetilde \w^{k} - \w^\ast), \label{eqn:p3}
\end{eqnarray}

Now we further take expectation with respect to the random variable $i_k$ in~(\ref{eqn:p344}). Since both $\w^{k}, \w^\ast$ are irrelevant to the choice of $i_k$, there is $\mathbb{E} \Delta_k^\top (\widetilde \w^{k} - \w^\ast) = (\mathbb{E} \Delta_k)^\top (\widetilde \w^{k} - \w^\ast) = 0$. The term $\| \Delta_k \|^2$ can be bounded as below:
\begin{eqnarray}
&& \mathbb{E} \| \Delta_k \|^2  \nonumber \\
&=& \mathbb{E} \| \nabla \psi_{i_k}({\w}^{k-1}) - [\nabla h_{i_k}(\w_s) - \nabla H(\w_s)] \nonumber \\
&& - \nabla P(\w^{k-1})\|^2 \nonumber \\
&=& \mathbb{E} \left\| \nabla \psi_{i_k}({\w}^{k-1}) - \nabla h_{i_k}(\w_s) \right\|^2 \nonumber \\
&& - \left\| \nabla H(\w_s) - \nabla P(\w^{k-1}) \right\|^2 \nonumber \\
&\le& \mathbb{E} \left\| \nabla \psi_{i_k}({\w}^{k-1}) - \nabla h_{i_k}(\w_s) \right\|^2 \nonumber \\
&\le& 2 \mathbb{E} \left\| \nabla \psi_{i_k}({\w}^{k-1}) - \nabla \psi_{i_k}({\w}^\ast) \right\|^2 \nonumber \\
&& + 2 \mathbb{E} \left\| \nabla h_{i_k}(\w_s) - \nabla \psi_{i_k}({\w}^\ast) \right\|^2\nonumber \\
&\le& 4 L ( F(\w^{k-1}) - F(\w^\ast) + F(\w_s) - F(\w^\ast) ) \nonumber \\
&& + 8 L \Delta_{H,P}, \nonumber
\end{eqnarray}
where we introduce the notation $\Delta_{H,P} = \max_{\w} | H(\w) - P(\w) |$ for the brevity of the presentation.

Therefore~(\ref{eqn:p3}) can be further transformed as
\begin{eqnarray}
&& \mathbb{E} \| \w^{k}-\w^\ast \|^2 \nonumber \\
&\le& \| \w^{k-1}-\w^\ast \|^2 -2 \eta ( \mathbb{E} F(\w^{k}) - F(\w^\ast) )\nonumber \\
&&+ 8 \eta^2 L ( F(\w^{k-1}) - F(\w^\ast) + F(\w_s) - F(\w^\ast) ) \nonumber \\
&& + 16 \eta^2 L \Delta_{H,P}. \label{eqn:p6}
\end{eqnarray}

After all $m$ iterations have been completed, we set $\w_{s+1} = \arg_{\w^k, k=1,\ldots,m} \min F(\w^k)$. Summing~(\ref{eqn:p6}) over $k = 1, \ldots, m$, we obtain
\begin{eqnarray}
&& \textstyle -\frac{2}{\mu_P + \mu_R} \left[ F(\w_s) - F(\w^\ast) \right] \nonumber \\
&\le& \mathbb{E} \| \w^{m}-\w^\ast \|^2 - \| \w^{0}-\w^\ast \|^2 \nonumber \\
&\le&   -2 \eta \sum_{k=1}^m \left( \mathbb{E} F(\w^{k}) - F(\w^\ast) \right)\nonumber \\
&&  + 8 \eta^2 L \sum_{k=1}^m \left( \mathbb{E} F(\w^{k-1}) - F(\w^\ast) \right) \nonumber \\
&&  + 8 \eta^2 L m (F(\w_s) - F(\w^\ast) + 16 \eta^2 L m \Delta_{H,P} \nonumber \\
&\le&  -2 \eta (1-4 \eta L) \sum_{k=1}^m \left( \mathbb{E} F(\w^{k}) - F(\w^\ast) \right)\nonumber \\
&&  + 8 \eta^2 L \left( F(\w^{0}) - F(\w^\ast) - (\mathbb{E} F(\w^{m}) - F(\w^\ast)) \right) \nonumber \\
&&  + 8 \eta^2 L m (F(\w_s) - F(\w^\ast) + 16 \eta^2 L m \Delta_{H,P} \nonumber \\
&\le&  -2 \eta (1-4 \eta L) m \left( F(\w_{s+1}) - F(\w^\ast) \right)\nonumber \\
&&  + 8 \eta^2 L (m+1) (F(\w_s) - F(\w^\ast) \nonumber \\
&& + 16 \eta^2 L m \Delta_{H,P}, \label{eqn:p7}
\end{eqnarray}
where the first inequality follows from the strong convexity of $F(\w)$ and the last inequality is derived from the construction of $\w_{s+1}$.

Re-arranging~(\ref{eqn:p7}) and using the notations $\rho,\Delta$ in~(\ref{eqn:main_1})(\ref{eqn:main_2}) respectively, we obtain
\begin{eqnarray}
F(\w_{s+1})  - F(\w^\ast) - \Delta \le \rho \left[ F(\w_s) - F(\w^\ast) -\Delta \right].
\end{eqnarray}

Recursively expanding the right hand side of above inequality obtains the main theoretic result in~(\ref{eqn:main_3}).
\end{proof}
\end{theorem}

\textbf{Remarks:} Similar to the original SVRG and its variant Prox-SVRG, our proposed S3GD also enjoys an exponential rate of convergence. The convergence guarantee S3GD is comparably weaker, since it is not guaranteed to converge to the exact globally-optimal solutions. Instead, the theoretic analysis in Theorem~\ref{thm:main} essentially states that, when the step size $\eta$ is sufficiently small, the function value-descending process from $F(\w_0)$ to $F(\w^\ast) + \Delta$ admits a geometric rate. However, the optimization after reaching $F(\w^\ast) + \Delta$ is beyond the scope of Theorem~\ref{thm:main}, which we will investigate through empirical study in the experimental section. In a word, our proposed S3GD avoids the intensive computation of full gradient by relaxing the optimality guarantees.

The quantity $\Delta$ is primarily determined by the residual between the smooth component $P(\w)$ of the composite objective function and its anchor-based approximation $H(\w)$. Note that in Eqn.~(\ref{eqn:main_2}), when $\eta$ is proportional to $1/L$ and $m$ is sufficiently large, the coefficient in front of $\Delta_{H,P}$ tends to be a constant, rather than increasing with respect to larger $m$. Moreover, the maximum value of the step size $\eta$ is slightly lower than that in Prox-SVRG ($\eta < 1/(8L)$ in S3GD and $\eta < 1/(4L)$ in Prox-SVRG).

To mitigate the effect of $\Delta$, one can either reduce the value of $\eta$ or switch to the normal Prox-SVRG procedure when the estimated parameter vector is close to the global optimum. In this paper we leave such investigation to the future work, faithfully reporting the performance of S3GD using constant step sizes in all experiments.


\section{Experiments}
\label{sec:exp}

This section reports the numerical studies between our proposed S3GD and other competing algorithms.

\begin{figure*}[t]
\begin{center}
   \includegraphics[width=0.19\linewidth]{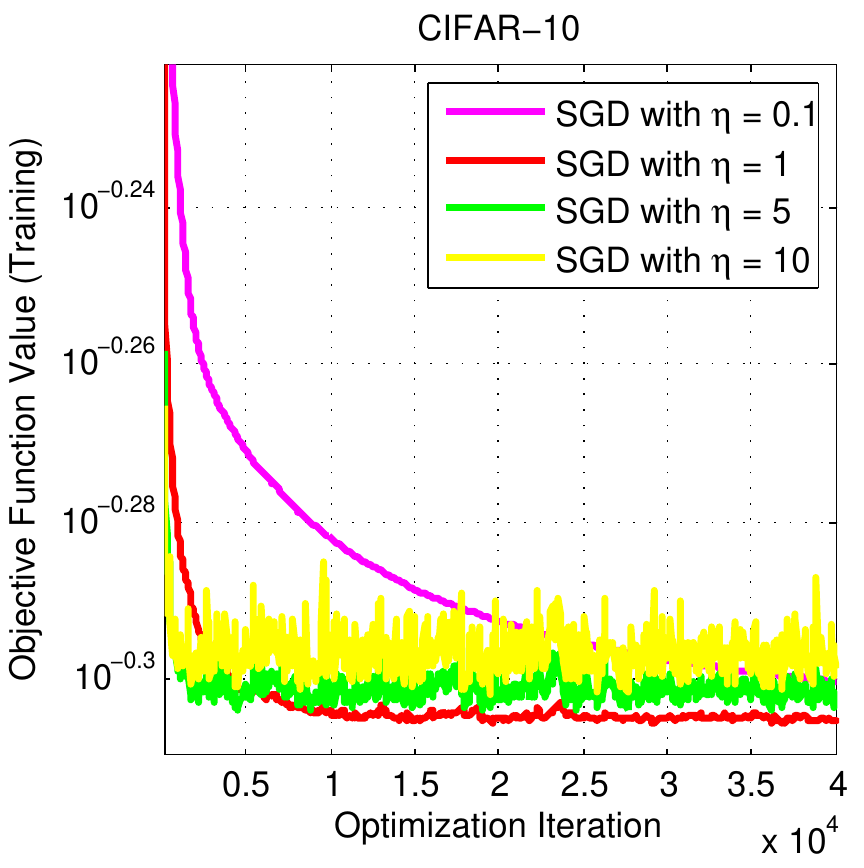}
   \includegraphics[width=0.19\linewidth]{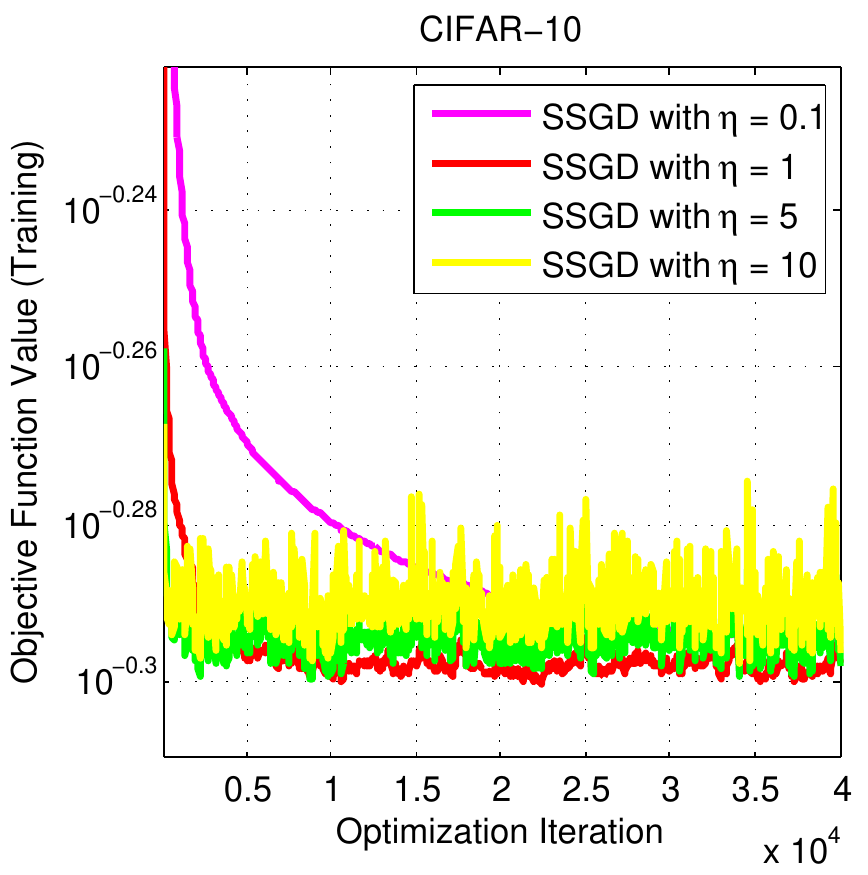}
   \includegraphics[width=0.19\linewidth]{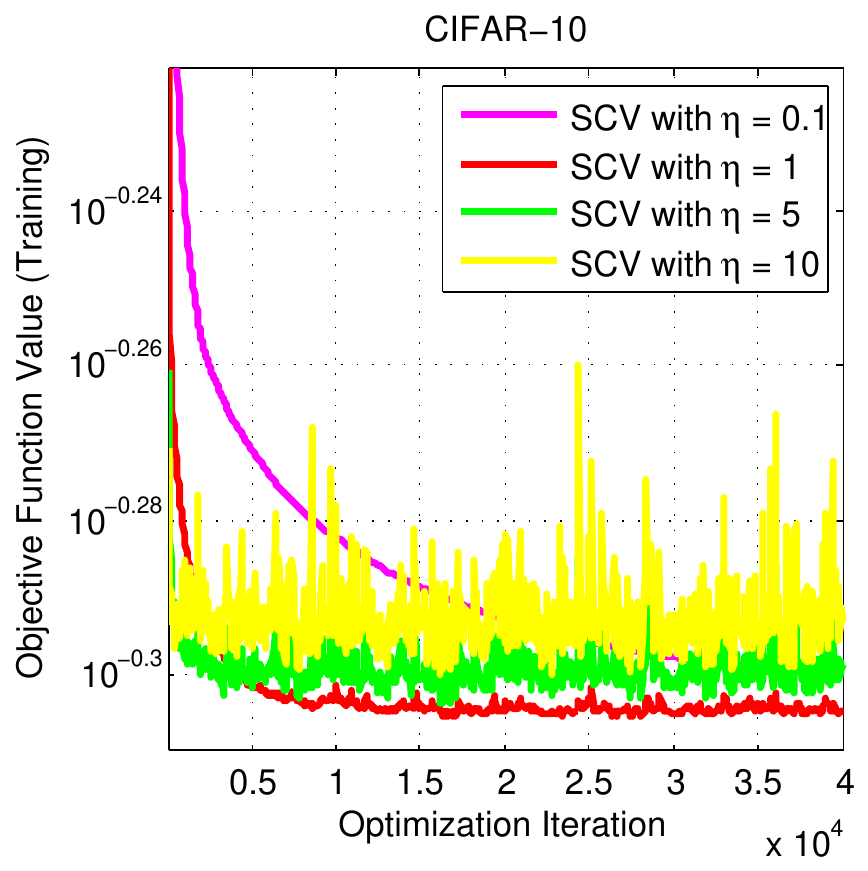}
   \includegraphics[width=0.19\linewidth]{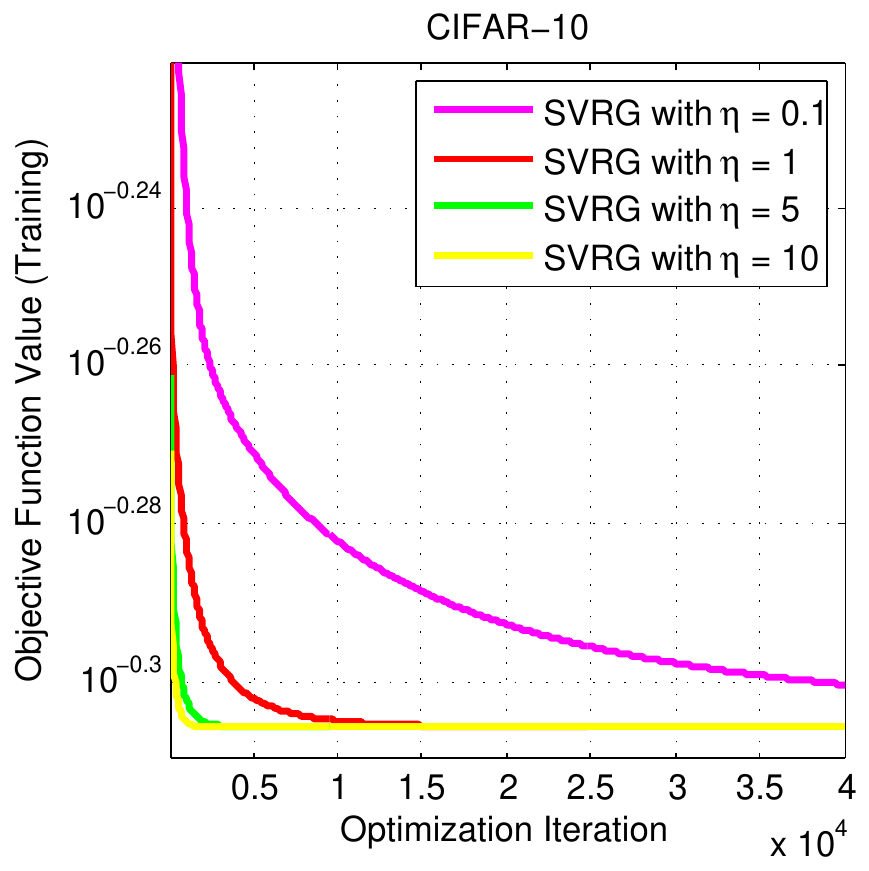}
   \includegraphics[width=0.19\linewidth]{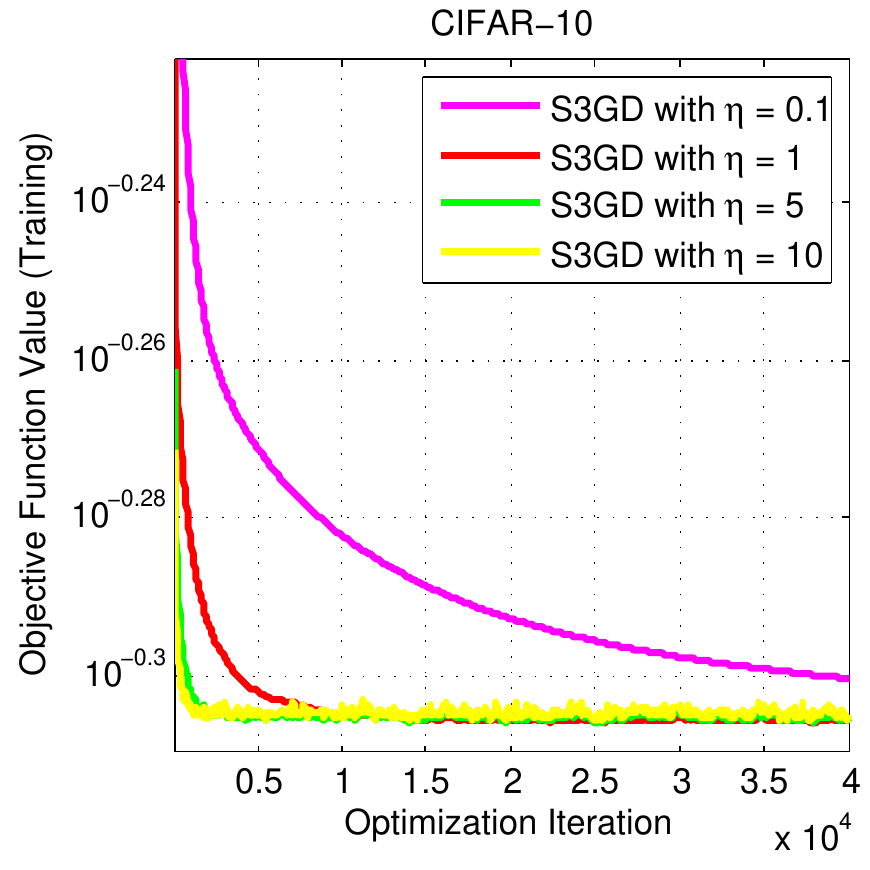}
\end{center}
   \caption{\small Investigation of the effect of step sizes on the convergence speed and solution stability. We select CIFAR10 as the testbed and report the training objective values under four different step size parameters. It is seen that large step sizes often indicate faster convergence yet also bring the risk of bouncing around the optimum. Variance reduction is thus critical for using large step sizes. Note that the objective values are plotted in logarithmic scale. Better viewing when enlarged and in color mode.}
\label{fig:stepsize}
\end{figure*}

\subsection{Description of Dataset and Applications}

To make the experiments comprehensive, we include nine benchmarks that cover a variety of heterogeneous tasks and different data scales: \emph{20-Newsgroups}\footnote{\url{http://qwone.com/~jason/20Newsgroups/}} which contains nicely-organized documents from 20 different news topics, \emph{WebSpam}\footnote{\url{http://www.csie.ntu.edu.tw/~cjlin/libsvmtools/datasets/binary.html}} represents a large collection of annotated spam or non-spam hosts labeled by a group of volunteers, \emph{IJCNN}\footnote{\url{http://www.geocities.com/ijcnn/nnc_ijcnn01.pdf}} for time-series data, \emph{KDD04\_bio} and \emph{KDD04\_phy}\footnote{\url{http://osmot.cs.cornell.edu/kddcup/datasets.html}} which correspond to the protein homology sub-task and quantum physics sub-task in KDD-Cup 2004 respectively,  \emph{covtype}\footnote{\url{https://archive.ics.uci.edu/ml/datasets/Covertype}} which includes cartographic variables for predicting forest cover type. We also include three computer vision benchmarks: \emph{CIFAR10}\footnote{\url{http://www.cs.toronto.edu/~kriz/cifar.html}} for image categorization, \emph{Kaggle-Face}\footnote{https://www.kaggle.com/c/challenges-in-representation-learning-facial-expression-recognition-challenge} for facial expression recognition and \emph{MED11}\footnote{\url{http://www.nist.gov/itl/iad/mig/med11.cfm}} for video event detection.

\begin{table}[t]
\centering
\begin{tabular}{lrrr}
  \hline\hline
 \textbf{Dataset} & \textbf{Train/Test Size} & \textbf{\#Dim} & \textbf{\#Class}   \\
  \hline
\textbf{20newsgroups} & 11,314 / 7,532 & 26,214 & 20   \\
\hline
\textbf{IJCNN} & 49,990 / 91,701 & 22 & 2  \\
\hline
 \textbf{WebSpam} & 280,000 / 70,000 & 254 & 2   \\
  \hline
\textbf{CIFAR10} & 50,000 / 10,000 & 800 & 10   \\
  \hline
  \textbf{Kaggle-Face} & 315,799 / 7,200 & 256 & 7 \\
 \hline
 \textbf{MED11} & 30,000 / 16,904 & 5,000 & 25 \\
 \hline
 \textbf{KDD04\_bio} & 120,000 / 25,751 & 74 & 2 \\
 \hline
 \textbf{KDD04\_phy} & 45,000 / 5,000 & 65 & 2 \\
 \hline
 \textbf{Covtype} & 500,000 / 81,012 & 54 & 2 \\
 \hline
\end{tabular}
\caption{\small Summary of the benchmarks used in the experiments. \textbf{\#Dim} and \textbf{\#Class} represent the number of feature dimensions and unique categories respectively.}
\label{table:data}
\end{table}

Table~\ref{table:data} summarizes the critical information for above-mentioned benchmarks. For most datasets, we adopt the defaulted train/test data split. Regarding the features, we either use the feature files provided by the benchmark organizers or extract them by ourselves. They may not necessarily bring state-of-the-art accuracy since our focus is investigating the convergence speed of the optimization methods instead of just driving for higher performance. The defaulted tasks defined on some benchmarks are multi-class classification. In these cases, a one-vs-rest scheme is applied to simplify the evaluations. We pick the category with the most training samples as the positive class and merge all rest categories as the negative class, converting it into a binary classification problem. Whenever the positive/negative data partitions are heavily unbalanced, we assign samples from positive/negative classes different weights such that the weight summarizations of the two classes are equal. More formally, let $\mathcal{Y}_+, \mathcal{Y}_-$ be the index sets of positive/negative classes respectively. The loss is calculated as
\begin{eqnarray}
P(\w) = \textstyle \frac{1}{|\mathcal{Y}_+|} \sum_{i \in \mathcal{Y}_+} \psi_i(\w) + \frac{1}{|\mathcal{Y}_-|} \sum_{i \in \mathcal{Y}_-} \psi_i(\w).
\label{eqn:pw}
\end{eqnarray}

In all experiments we stick to using the logistic loss function and Tikhonov regularization owing to their empirical popularity and non-linear property.

\begin{figure*}[t!]
\begin{center}
   \includegraphics[width=0.9\linewidth]{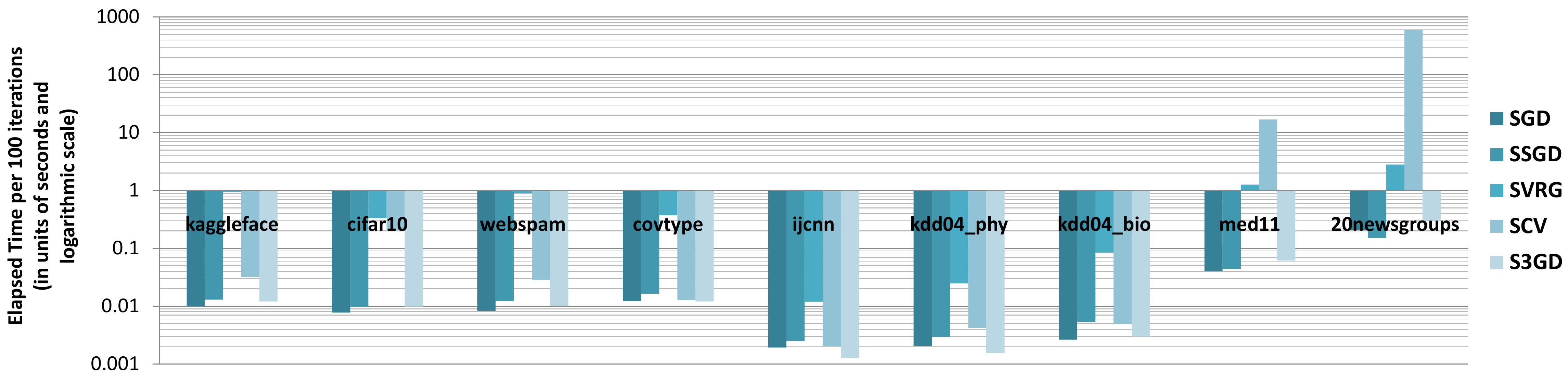}
\end{center}
   \caption{\small Iteration time complexities in terms of CPU times on all datasets. The time is recorded in seconds. To highlight subtle difference, we adopt logarithmic scale for the vertical axis. The length of time is visualized by a bar that points to its due value. See text for more explanations.}
\label{fig:time}
\end{figure*}

\subsection{Baseline Algorithms}

We make comparisons between the proposed S3GD and other four competitors, including
\begin{itemize}
\item \textbf{Mini-Batch Stochastic Gradient Descent (SGD)}: it represents the standard stochastic gradient method. At each iteration, the SGD algorithm randomly draws $p$ samples from the training set according to weight distribution specified in Eqn.~(\ref{eqn:pw}), calculate their respective stochastic gradient, and uniformly average these stochastic gradients.

\item \textbf{Stratified SGD (SSGD)}~\cite{ZhaoZ14b}: This method aims to improve the standard mini-batch SGD using data clustering and stratified sampling. SSGD ensures that each iteration will draw at least one sample from each data cluster (stratum). The inclusion is to contrast different ways of using global information about data. For fair comparison we set the number of clusters to be $p$.

\item \textbf{Stochastic Variance Reduction Gradient (SVRG)}: This original idea work of SVRG is found in~\cite{Johnson013}. However, it does not handle non-smooth functions. In the comparison we adopt the extension proposed in~\cite{Xiao014}. Inheriting the two nested loops of SVRG, one of the key parameters in~\cite{Xiao014} is the the maximal iteration number in the inner loop. The authors suggest this parameter shall be sufficiently large for achieving better loss bound. We fix this parameter to be always 50 in all experiments, which empirically provides a good balance between convergence speed and heavy complexity caused by exact gradient estimation.

\item \textbf{Stochastic Control Variate (SCV)}~\cite{WangCSX13}: This is another semi-stochastic gradient method that reports state-of-the-art speed and accuracies. The method relies on the utilization of data statistics such as low-order moments to define ``control variate". The authors rigourously prove the reduction in noisy gradient variance under mild assumptions. Note that for features in high dimension, the computation of data statistics can be its computational bottleneck.

\end{itemize}

\begin{figure*}[t]
\begin{center}
\includegraphics[height=0.24\linewidth,width=0.32\linewidth,keepaspectratio=false]{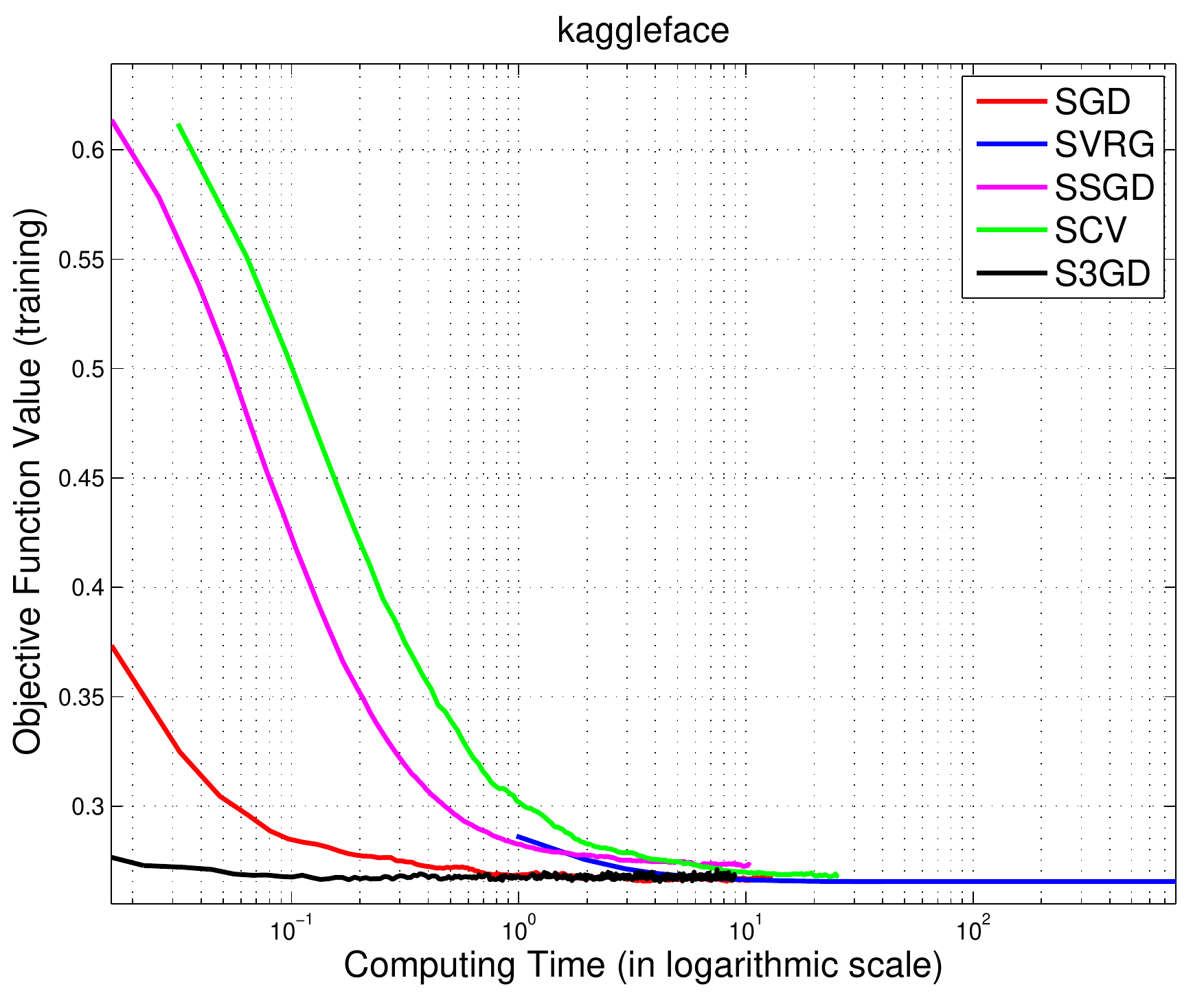}
\includegraphics[height=0.24\linewidth,width=0.32\linewidth,keepaspectratio=false]{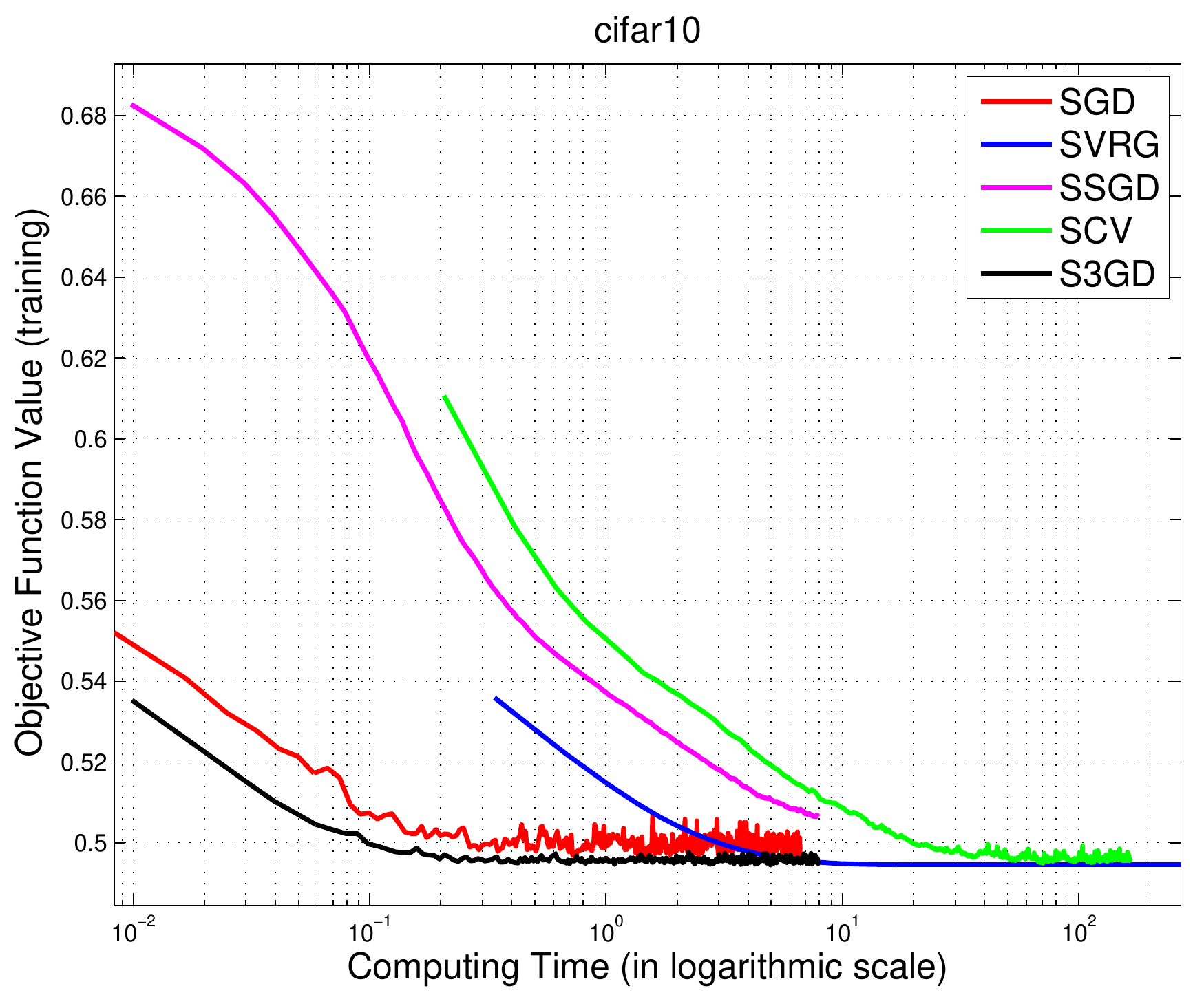}
\includegraphics[height=0.24\linewidth,width=0.32\linewidth,keepaspectratio=false]{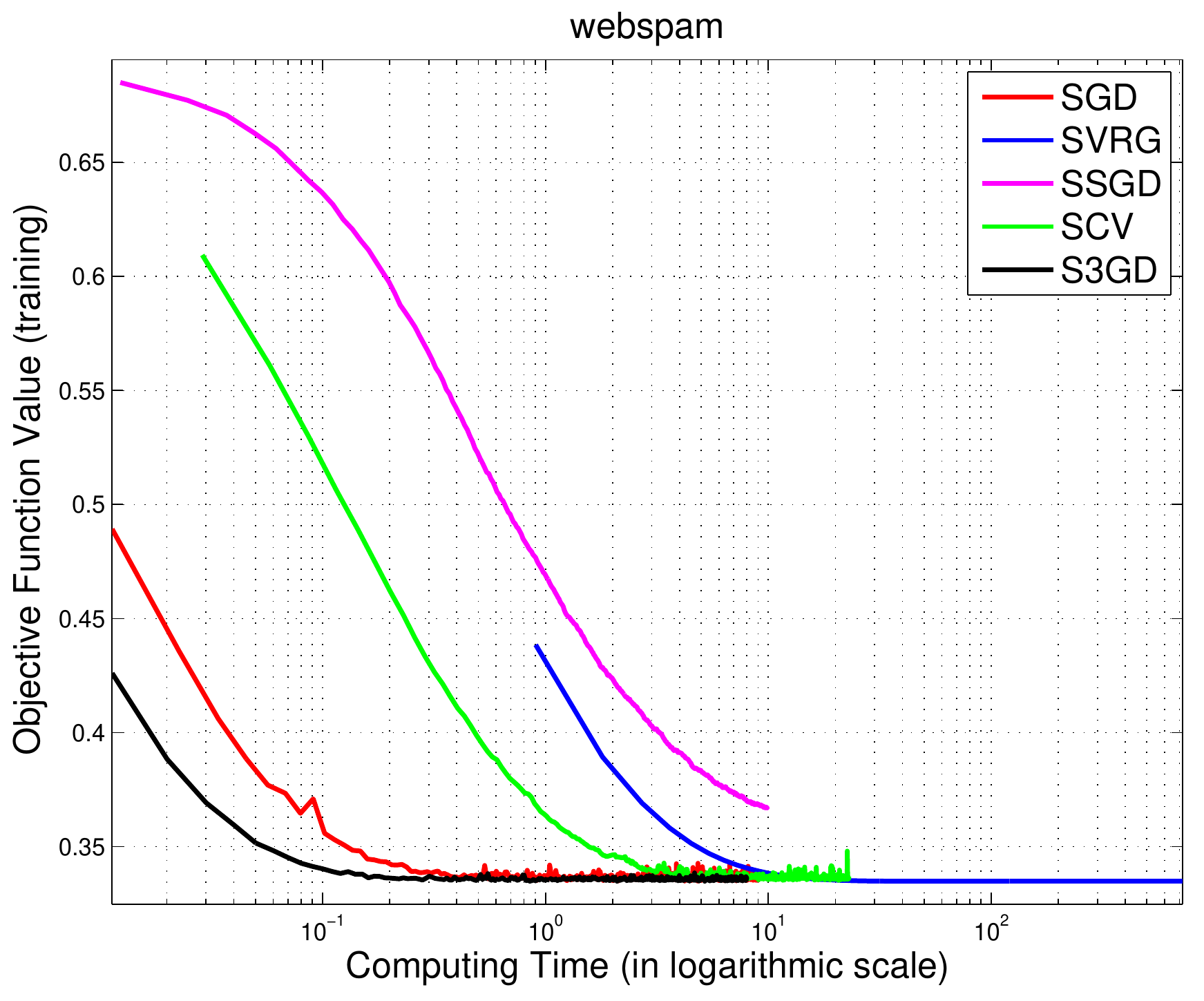}
\includegraphics[height=0.24\linewidth,width=0.32\linewidth,keepaspectratio=false]{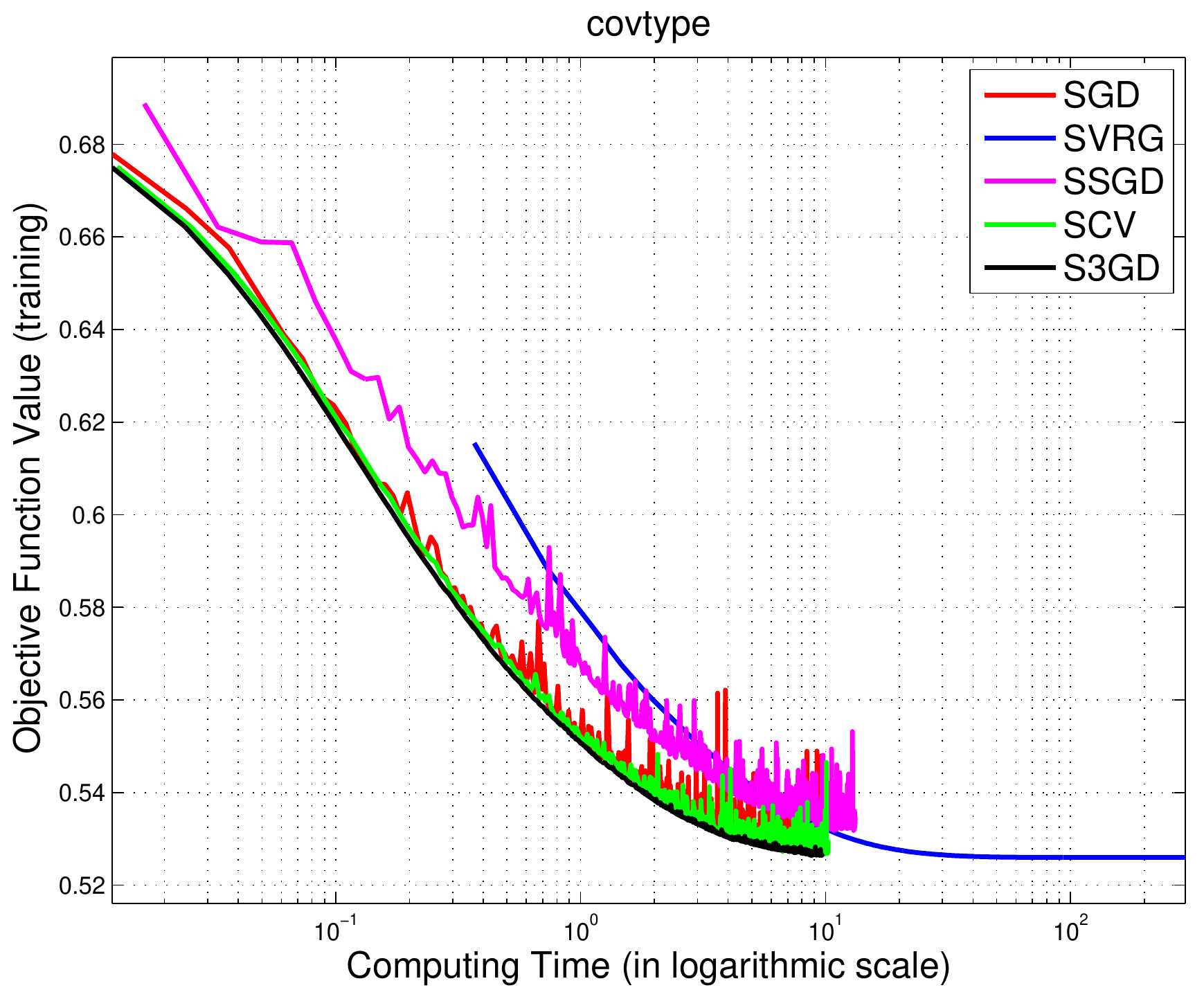}
\includegraphics[height=0.24\linewidth,width=0.32\linewidth,keepaspectratio=false]{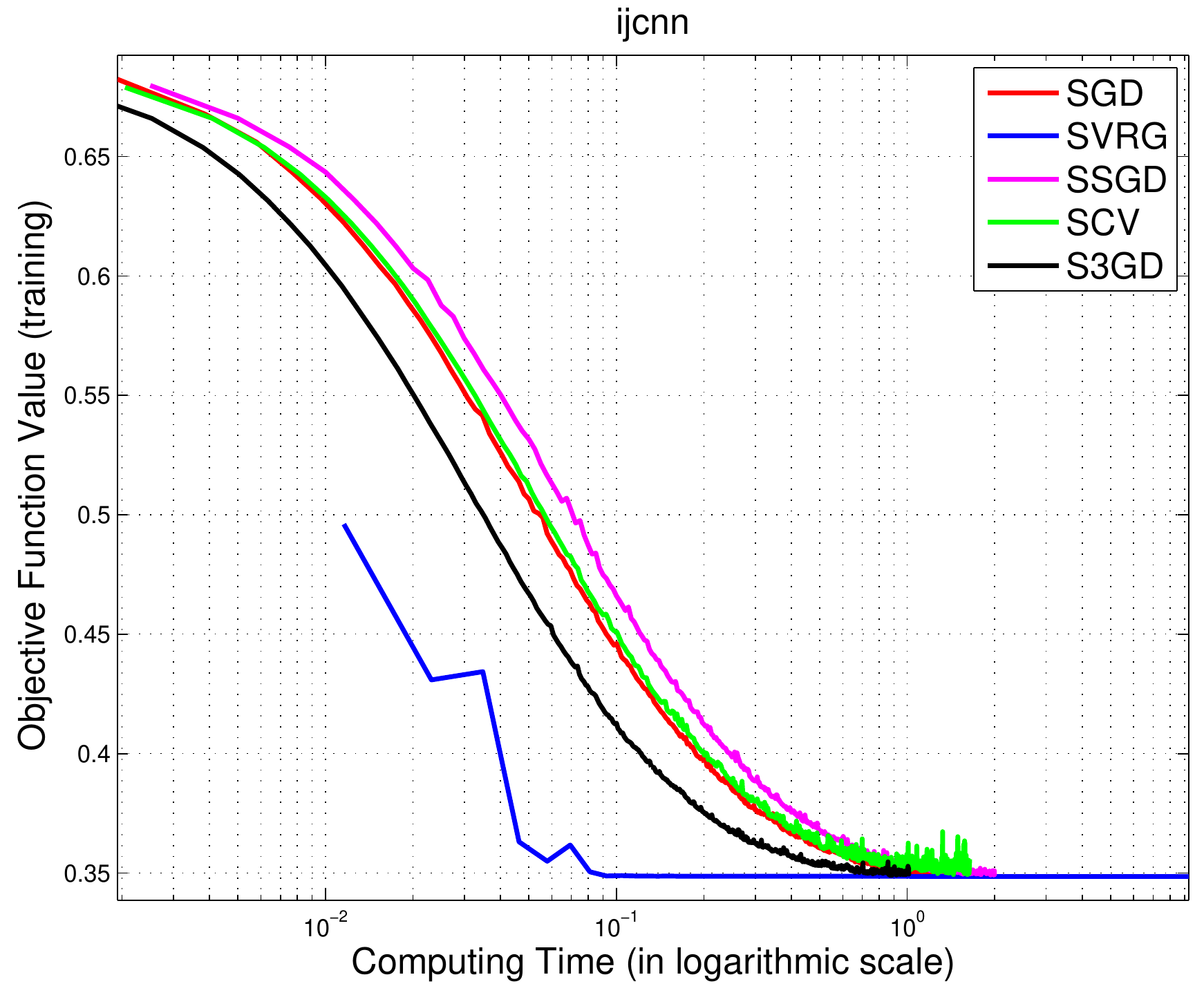}
\includegraphics[height=0.24\linewidth,width=0.32\linewidth,keepaspectratio=false]{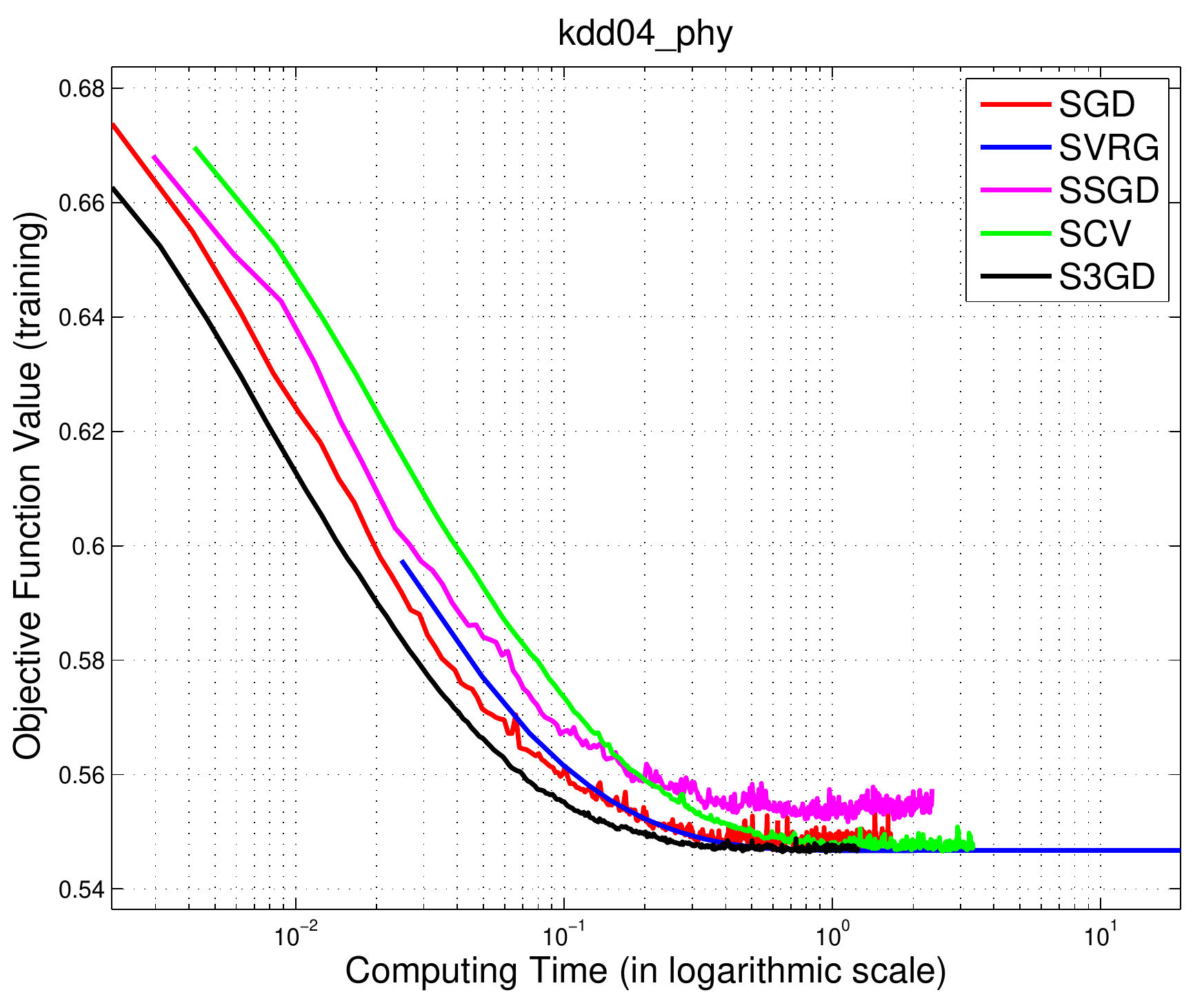}
\includegraphics[height=0.24\linewidth,width=0.32\linewidth,keepaspectratio=false]{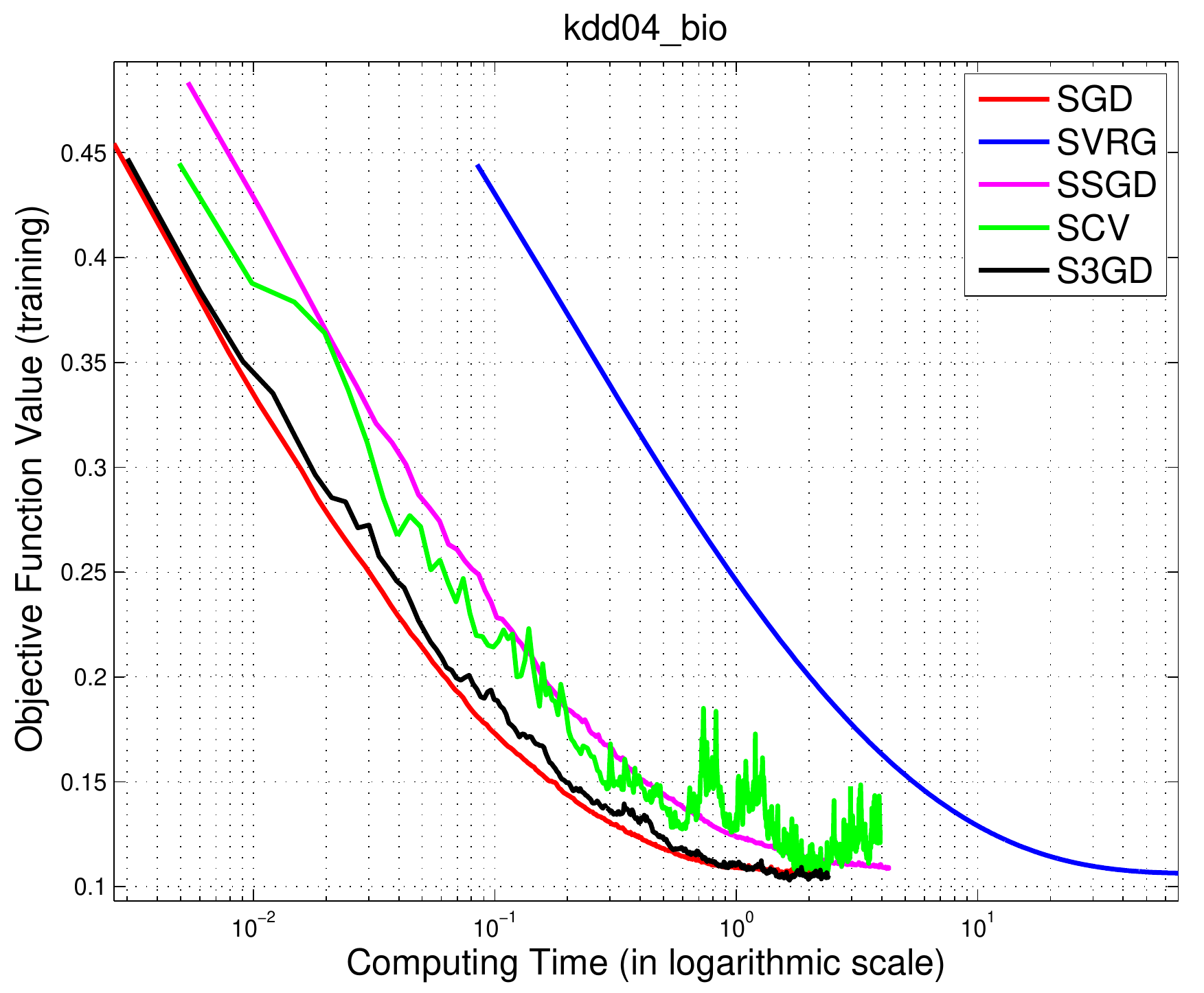}
\includegraphics[height=0.24\linewidth,width=0.32\linewidth,keepaspectratio=false]{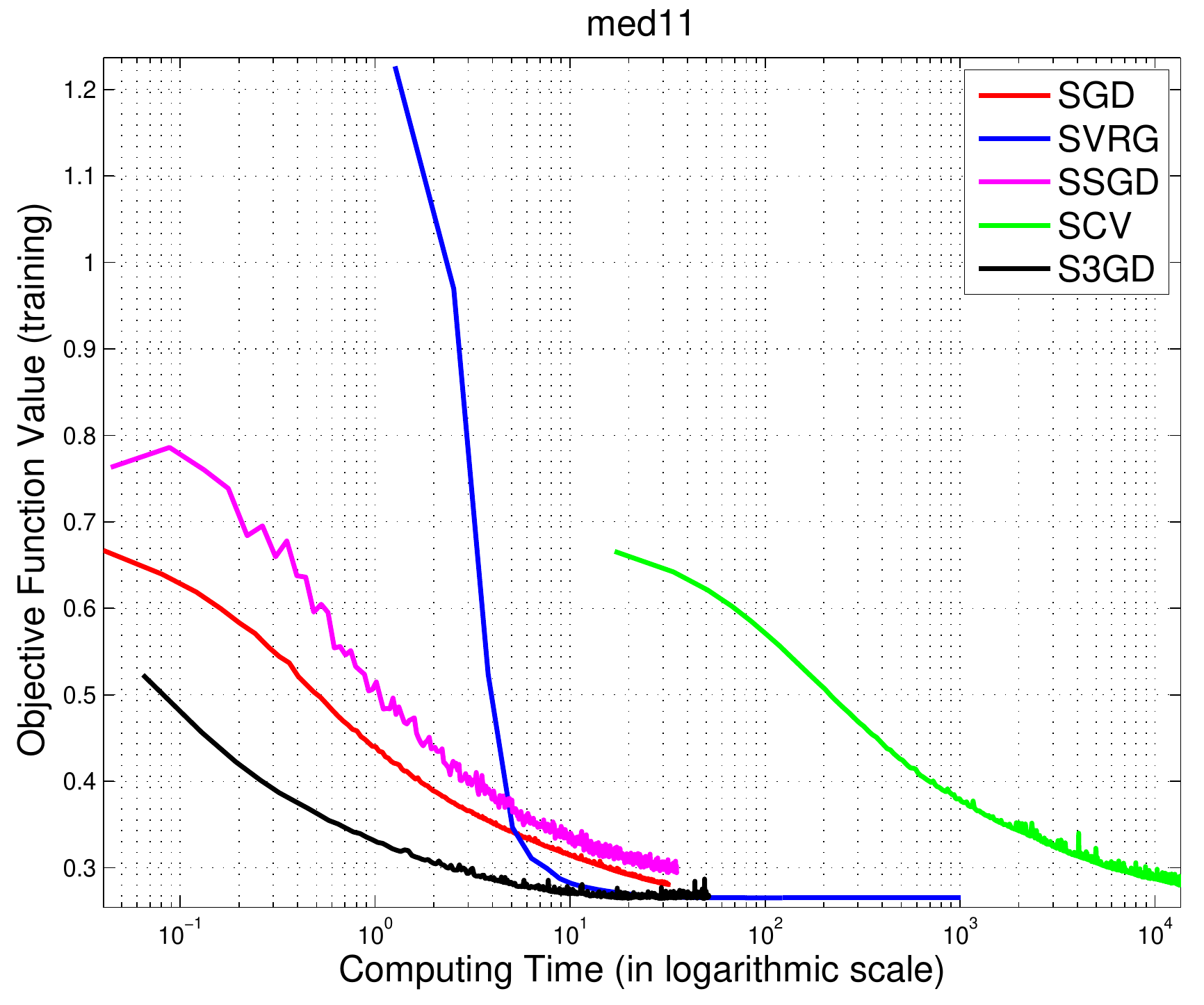}
\includegraphics[height=0.24\linewidth,width=0.32\linewidth,keepaspectratio=false]{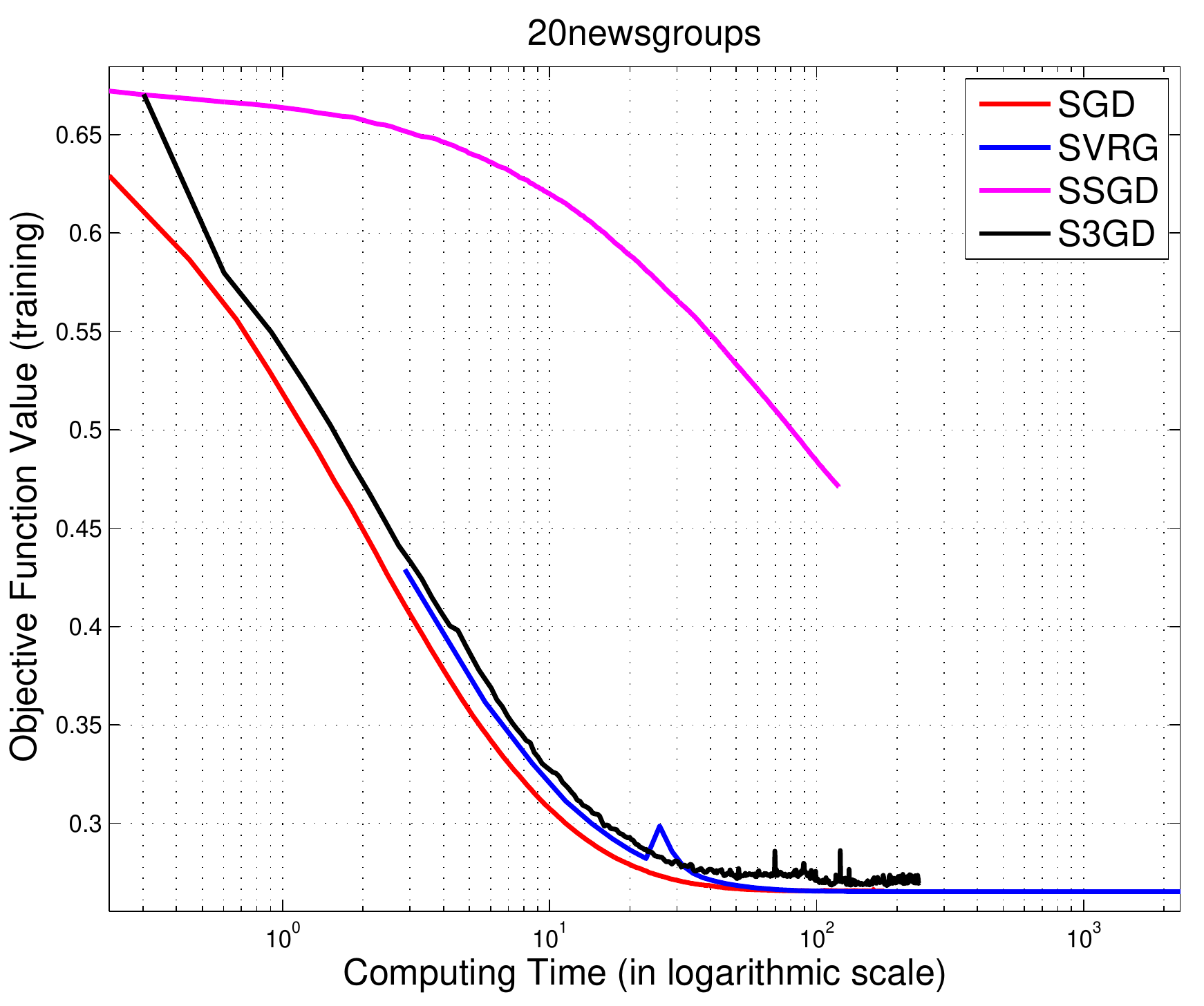}
\end{center}
   \caption{\small Training objective values for all referred algorithms in this paper on 9 datasets. The horizontal axis correspond to CPU times (in seconds) in the logarithmic scale. Note that we do not report the performance of SCV on 20newsgroups since it takes five days for accomplishing all 40,000 iterations. Better viewing in color.}
\label{fig:1}
\end{figure*}

\begin{figure*}[t]
\begin{center}
\includegraphics[height=0.24\linewidth,width=0.32\linewidth,keepaspectratio=false]{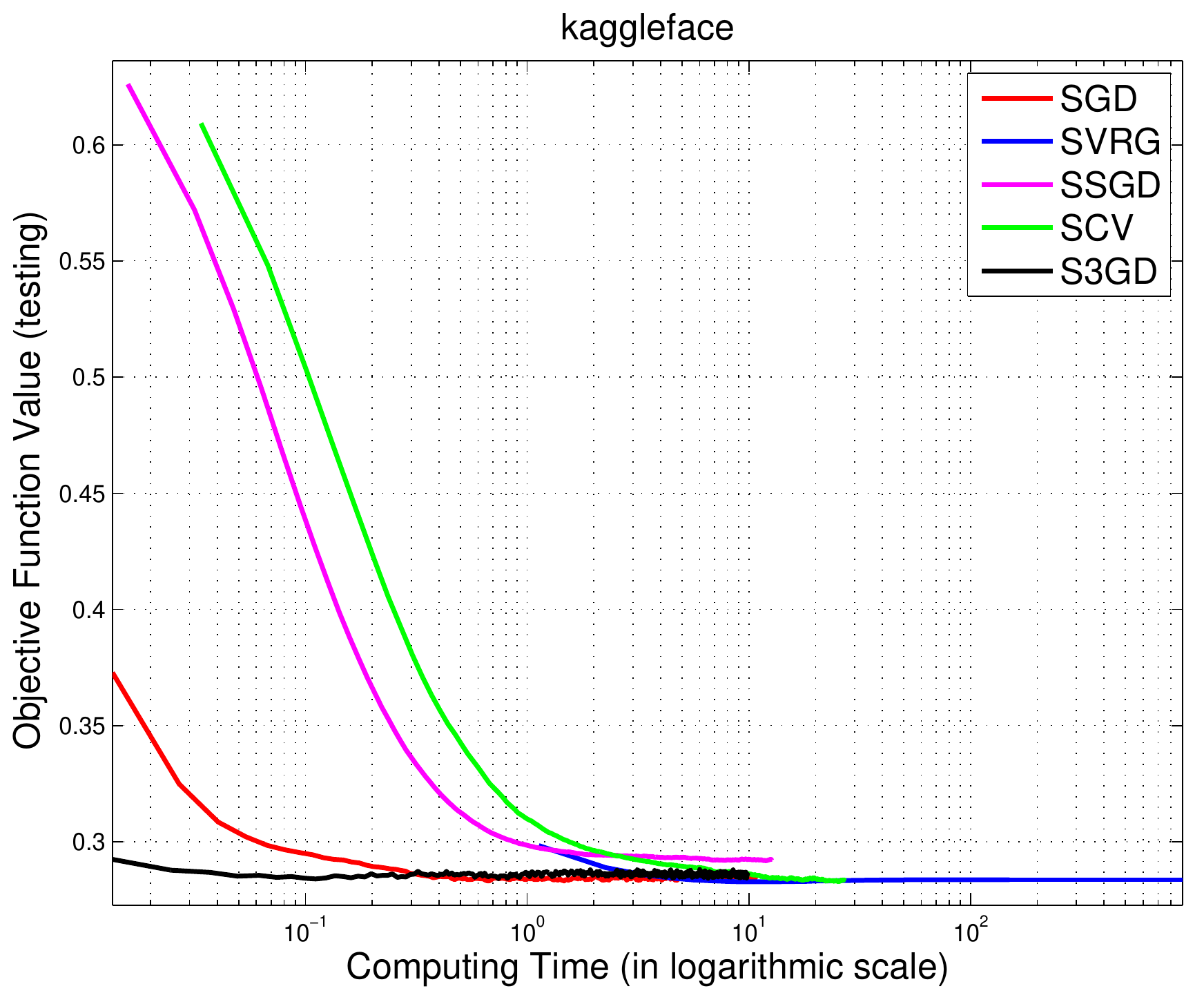}
\includegraphics[height=0.24\linewidth,width=0.32\linewidth,keepaspectratio=false]{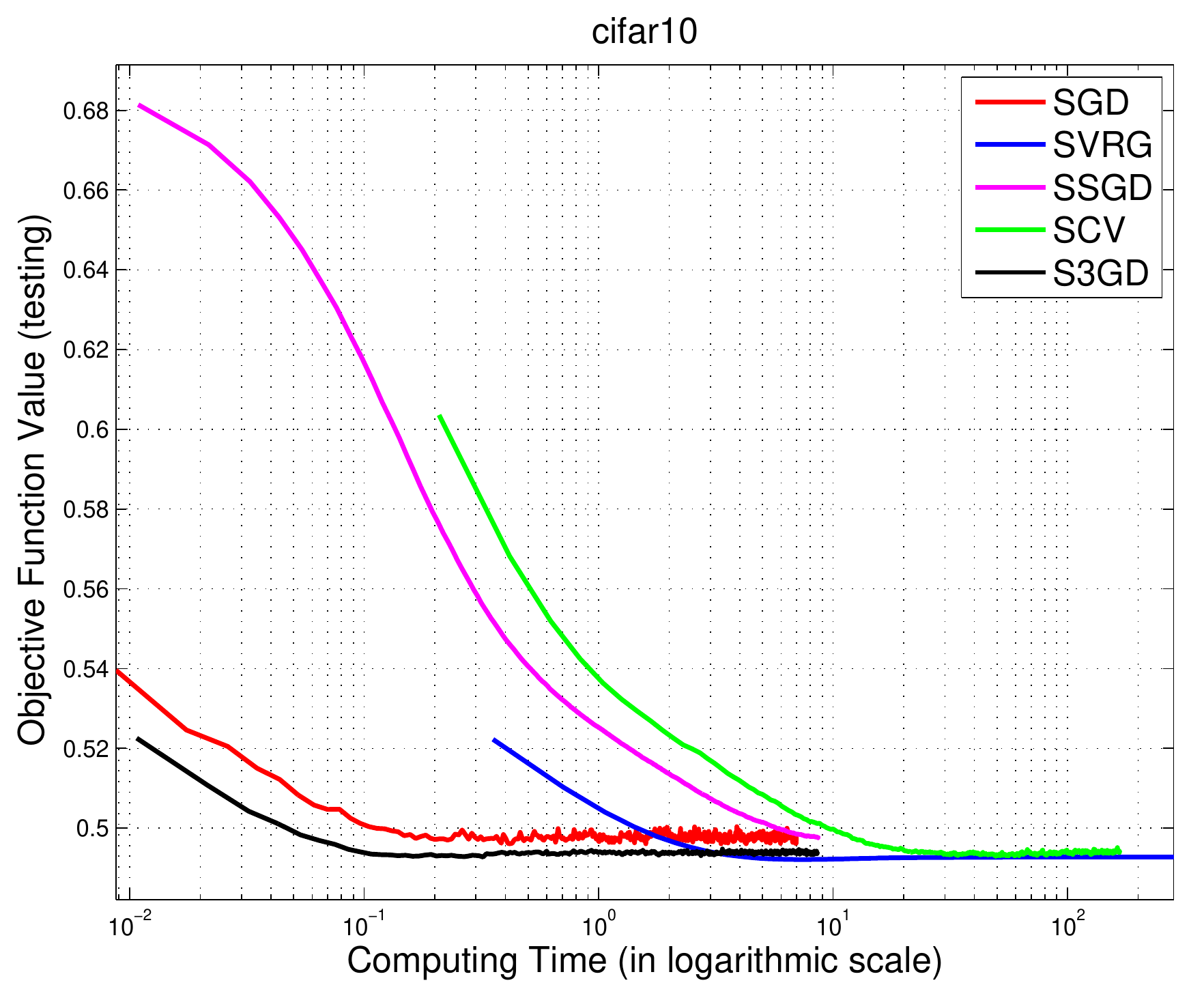}
\includegraphics[height=0.24\linewidth,width=0.32\linewidth,keepaspectratio=false]{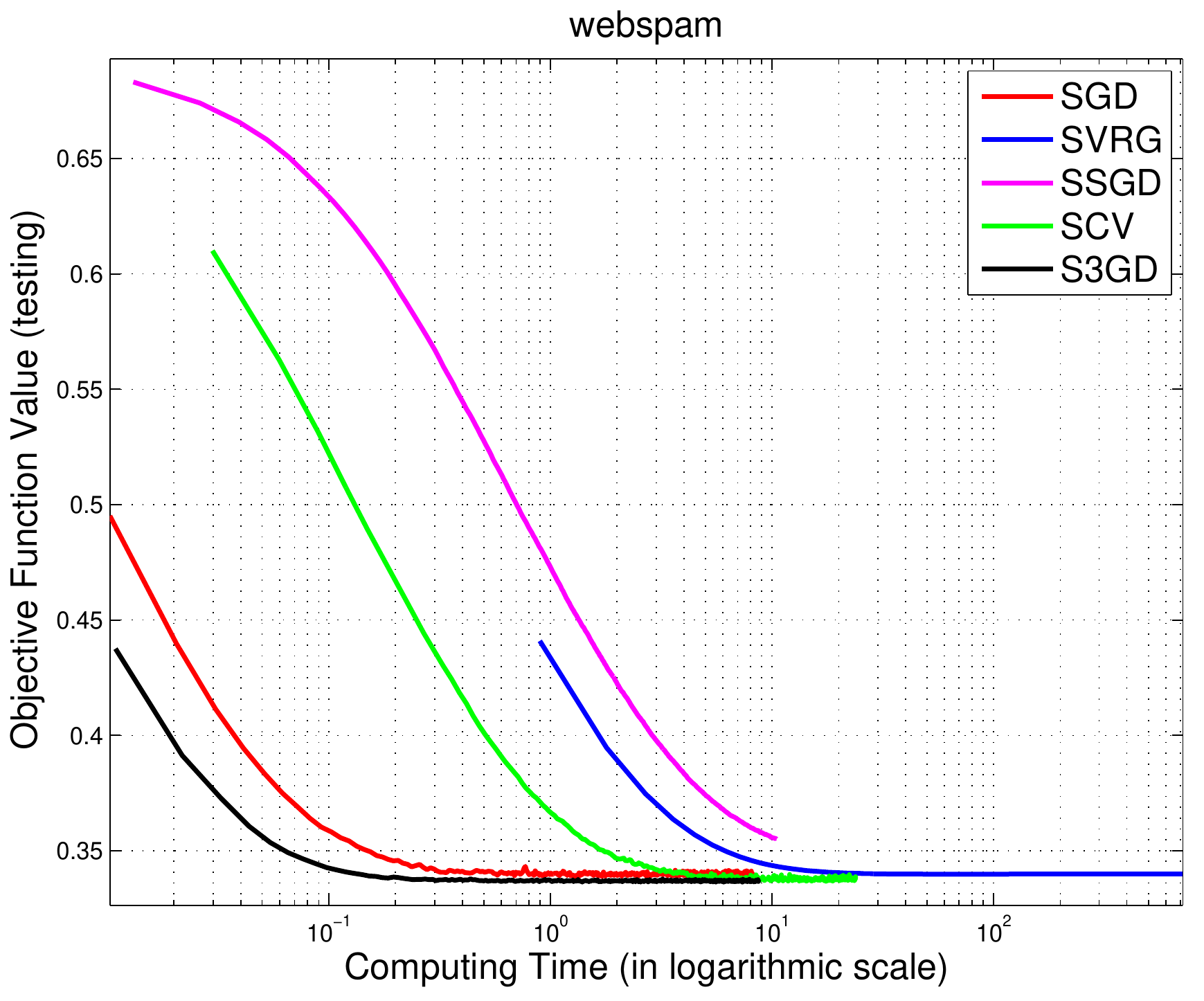}
\includegraphics[height=0.24\linewidth,width=0.32\linewidth,keepaspectratio=false]{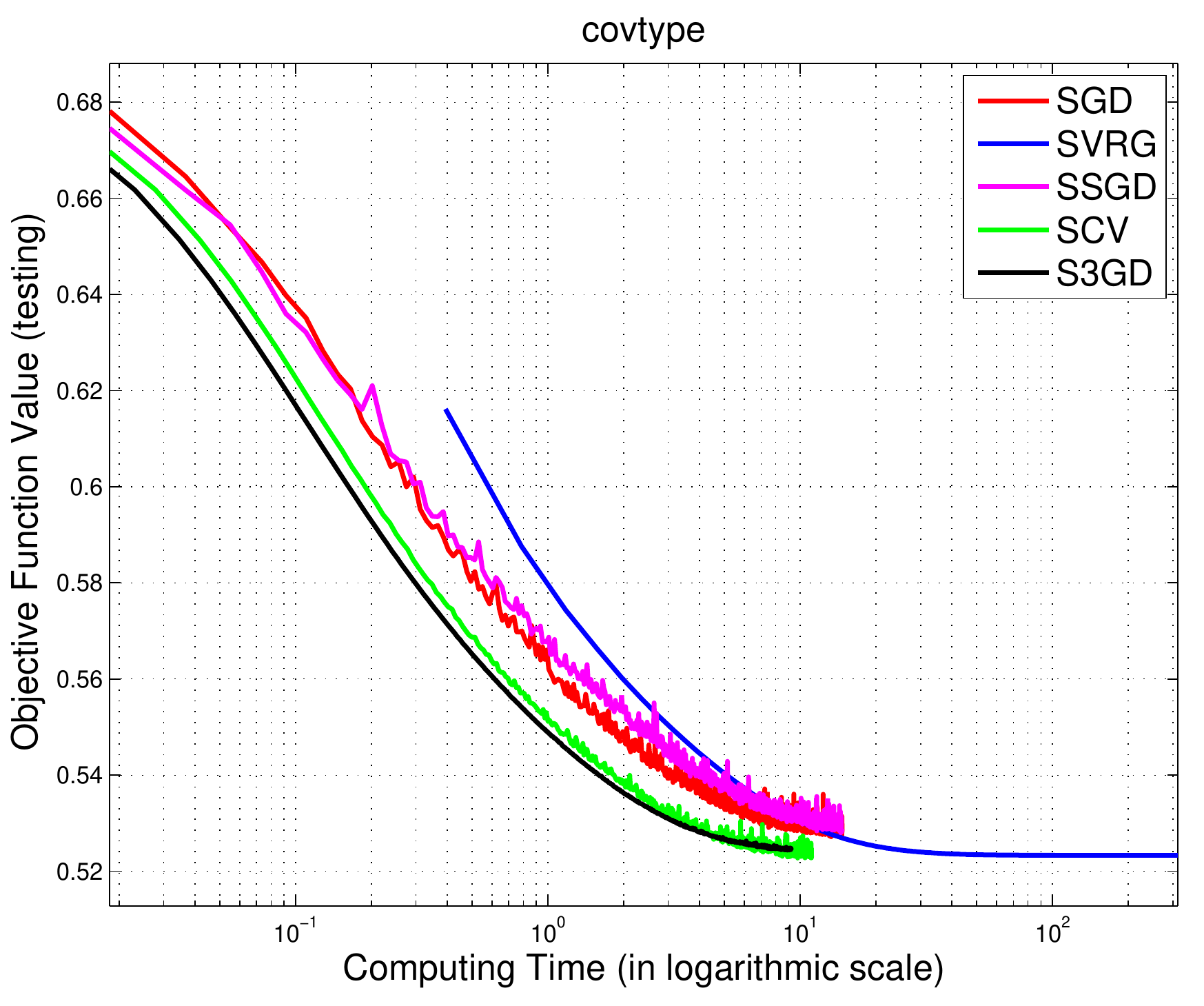}
\includegraphics[height=0.24\linewidth,width=0.32\linewidth,keepaspectratio=false]{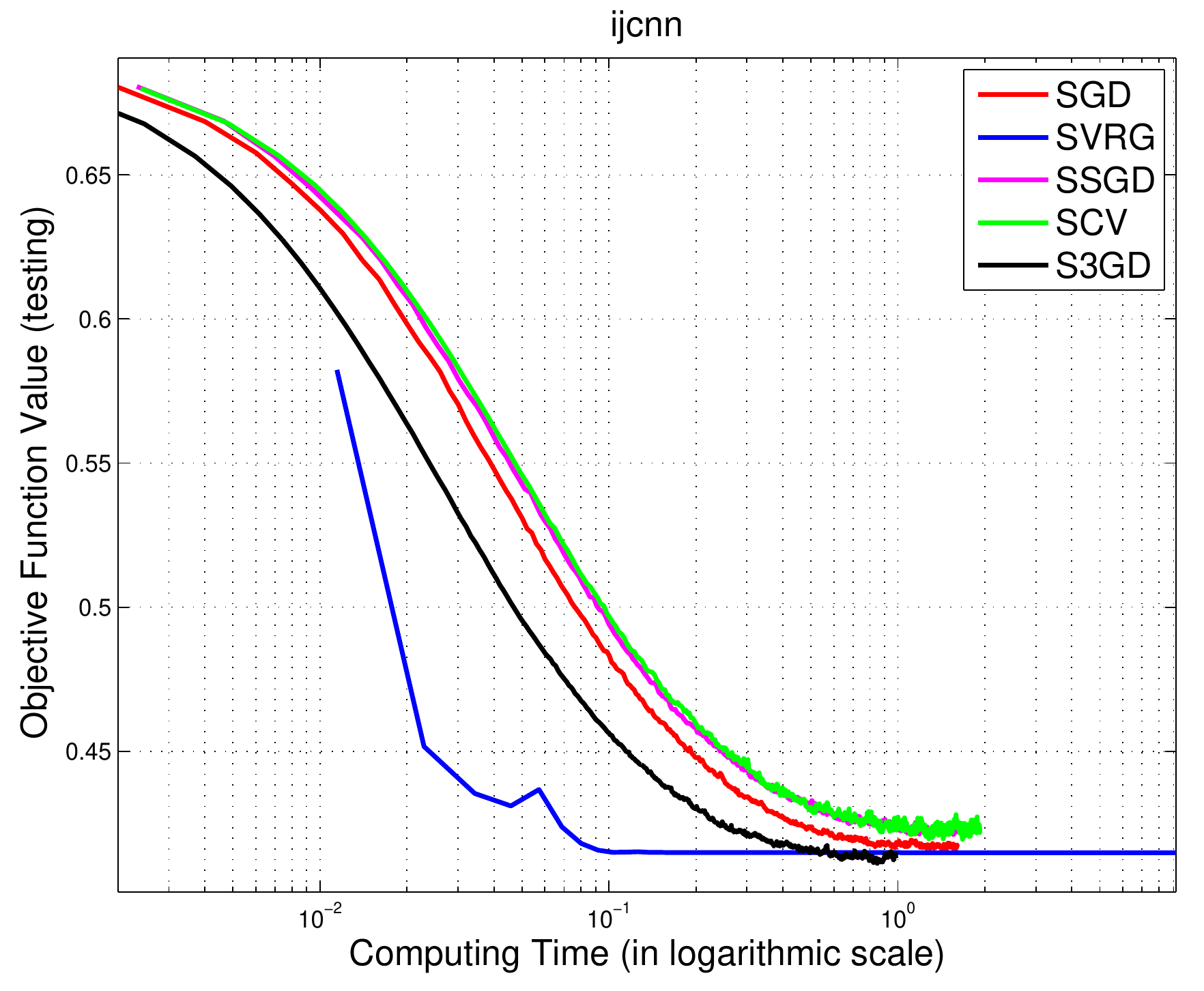}
\includegraphics[height=0.24\linewidth,width=0.32\linewidth,keepaspectratio=false]{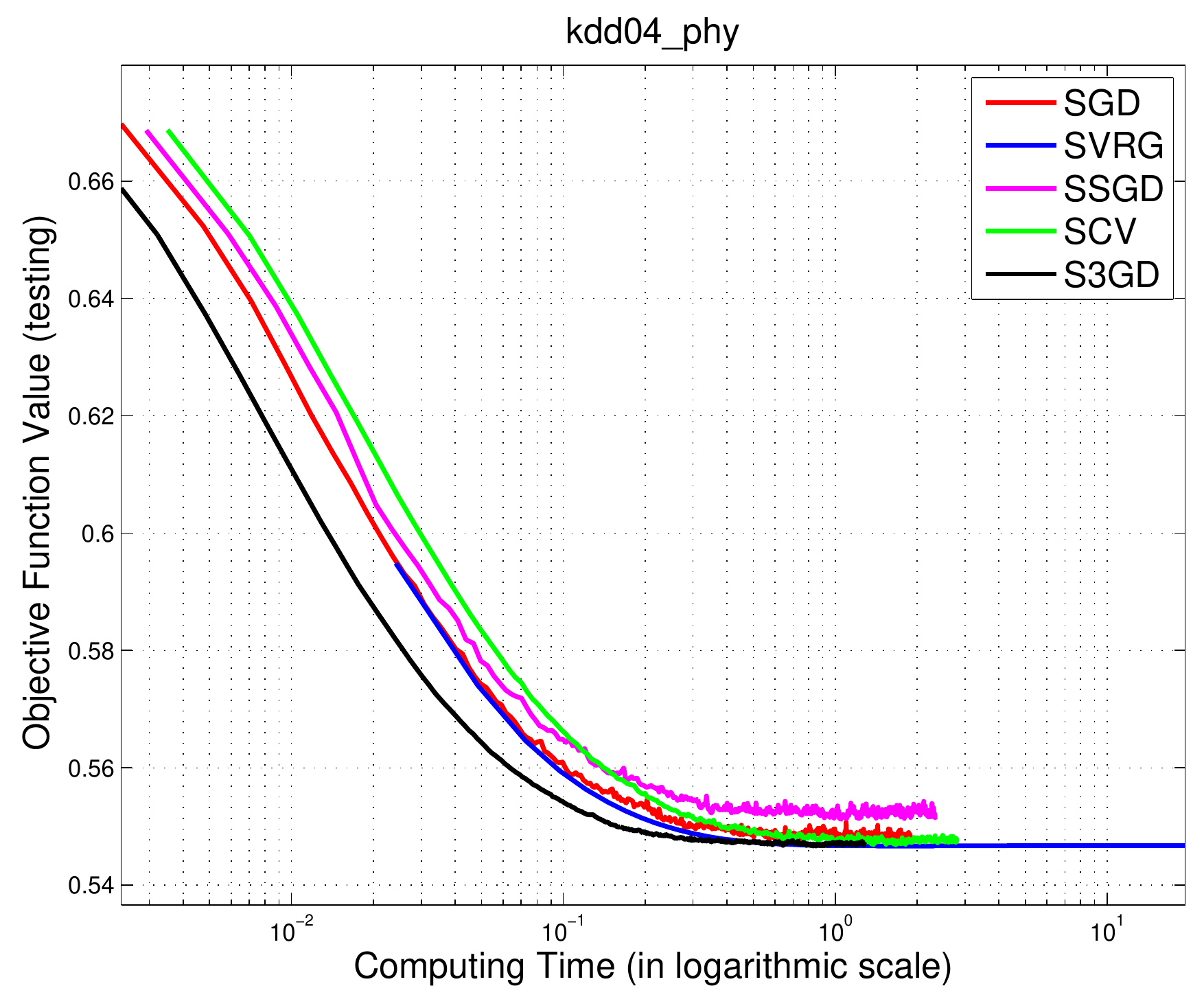}
\includegraphics[height=0.24\linewidth,width=0.32\linewidth,keepaspectratio=false]{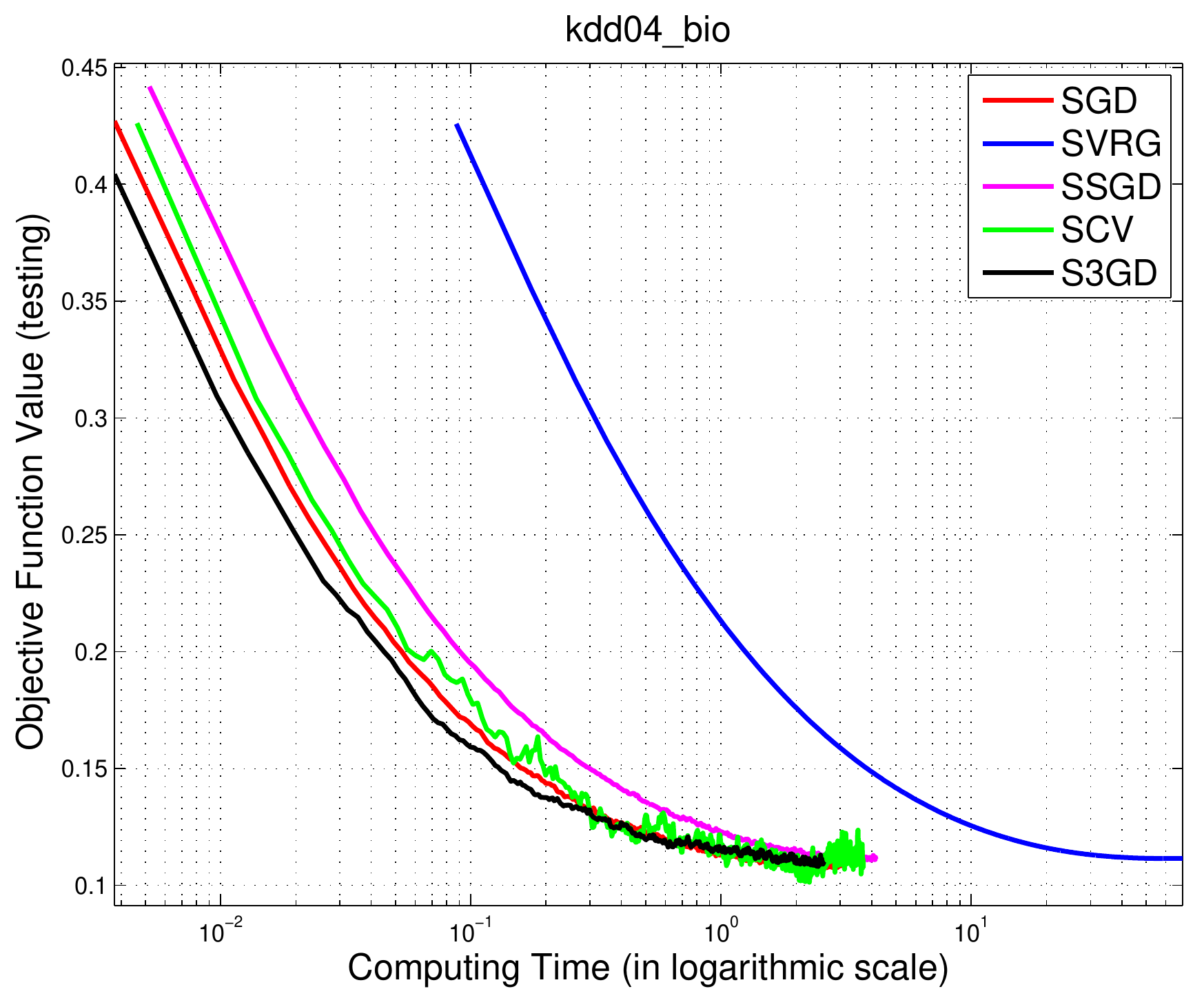}
\includegraphics[height=0.24\linewidth,width=0.32\linewidth,keepaspectratio=false]{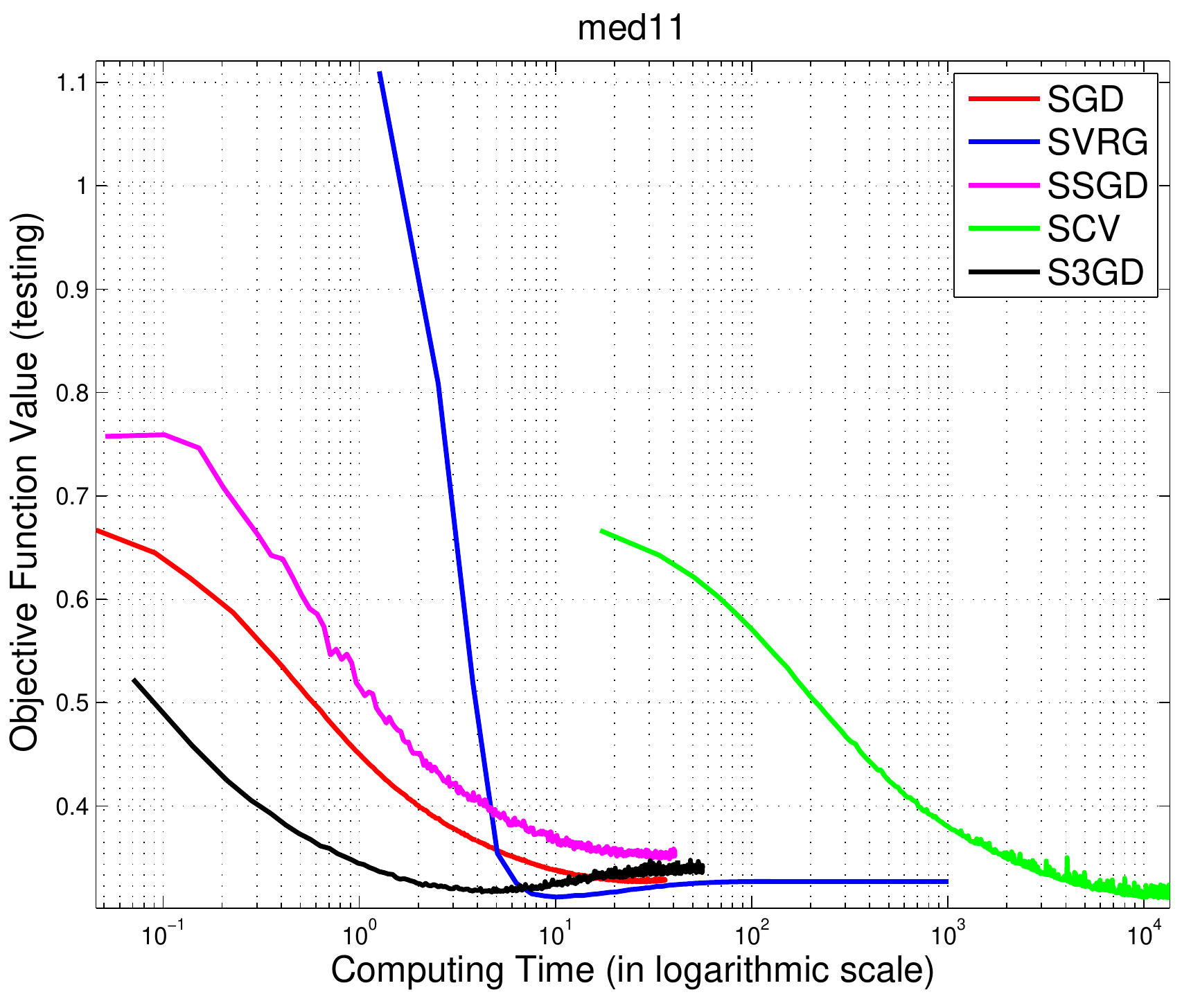}
\includegraphics[height=0.24\linewidth,width=0.32\linewidth,keepaspectratio=false]{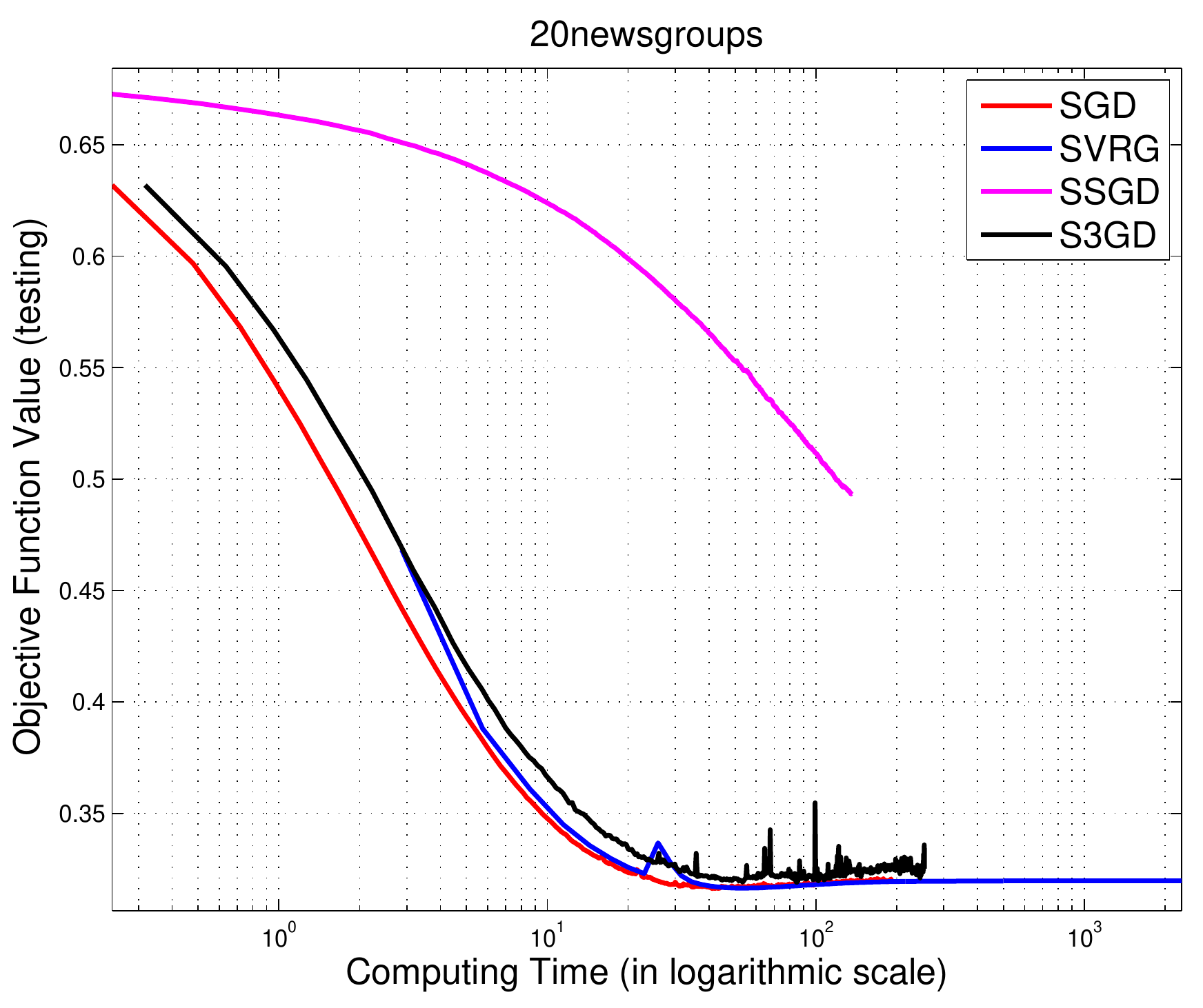}
\end{center}
   \caption{\small Testing objective values for all referred algorithms in this paper on 9 datasets. }
\label{fig:1b}
\end{figure*}

\subsection{Evaluation Settings}

For all experiments, we fix the parameter $\lambda = 10^{-3}$ for the Tikhonov regularization. The maximal iteration parameter $k_{in}$ in the inner loop of S3GD is fixed to be 20. Each mini-batch contains $p=10$ random samples. For S3GD, $m=100$ anchors are generated on all datasets. We implemented all baseline algorithms and S3GD in optimized C++ programs. The experiments are conducted on shared servers in an industrial research lab. Each of the servers is equipped with 48 CPU cores and 400GB physical memory. Five independent trials are performed for all algorithms and the averaged results are reported. The entire experiments take about one day on five servers.

There are two indices which are utterly crucial for evaluating a gradient based optimization scheme: the correlation (or variance) between (semi)stochastic gradient and exact gradient, and the maximal step size which ensures the stability of the optimization. In the literature of stochastic gradient methods, both decayed and constant step sizes are widely adopted. We find that tuning the decayed step sizes is very tricky, which makes a fair comparison among different algorithms difficult. Therefore we focus on the results using constant step sizes.

Large step sizes are always favored in practice since they expect improved convergence speed. To see this point, in Figure~\ref{fig:stepsize} we plot the objective values in each iteration of the training stage on CIFAR10. For all baseline algorithms and our proposed S3GD, the convergence curves under four different constant step sizes $\eta = \{0.1, 1, 5, 10\}$ are recorded and plotted. Obviously SVRG and S3GD are two most stable algorithms even operating with large step size parameters. All other three algorithms drastically fluctuate when their current solutions approach the global optimum, even with the moderate parameter $\eta = 5$.
This empirical investigation highlights the importance of choosing proper step size.

To fairly compare different algorithms, we evaluate them under the parameter set $\eta \in \{0.1, 1, 5, 10\}$ and report the performance with the largest step size that satisfies the following stability condition:
\begin{eqnarray}
\eta^\ast = \max ~\eta \quad s.t.~~F(\w; \eta)    \le (1 + \epsilon) F(\w^\ast), \label{eqn:eta}
\end{eqnarray}
where $F(\w^\ast)$ denotes the objective value at the global optimum $\w^\ast$. $F(\w; \eta)$ is the converged point using step size $\eta$. In the experiments we average the last 5,000 optimization iterations to obtain $F(\w; \eta)$. $\epsilon$ is set to be 0.01 in all cases. The condition aims to abandon any step size parameter that drives the solution crazily bounce around the global optimum.

Most of prior works~\cite{Johnson013,WangCSX13} report the performance with respect to iteration counts. We here argue that the evaluation shall take the iteration complexity into account. Recall that Table~\ref{table:complexity} summarizes the time and space iteration complexities for all algorithms. Importantly, the complexities of SVRG and SCV are dominated by the exact gradient computation and class-specific covariance matrix estimation. Both of them are expected to take longer time for accomplishing each iteration. Figure~\ref{fig:time} reports the time for performing 50 gradient descent iterations for all datasets and algorithms. The computing time is obtained by averaging all trials. It is observed that on most datasets, the standard SGD consumes the least time. SSGD and our proposed S3GD use slightly more time compared to SGD. The CPU time of SVRG and SCV are significantly larger. Specifically, SVRG is especially slow in comparison when facing large scale data set (such as Kaggle-face and covtype) and high feature dimensions (\emph{e.g.}, 5,000-dimensional features for MED11 and 26,214-dimensional features for 20newsgroups). Likewise, SCV is particularly slow when handling high-dimensional features. On the 20newsgroups, SCV requires 594 seconds for every 50 iterations, which is beyond the scope of most practitioners. In contrast, SGD and S3GD only need 0.21 and 0.30 seconds respectively. Therefore for fairness in comparison, we will majorly concern the performance with respect to CPU times.

\subsection{Quantitative Investigations}

\vspace{0.07in}
\noindent \textbf{Convergence Speed}: Figure~\ref{fig:1} shows the training objective values with respect to the CPU times. For all algorithms, the step-size parameters are chosen according to the criterion in~(\ref{eqn:eta}). Interestingly, though semi-stochastic gradient methods are proved to enjoy faster asymptotical convergence speed, most of them are not as ``economic" as standard SGD due to significantly higher iteration complexity. Our proposed S3GD exceptionally outperforms all other algorithms on 6 out of 9 datasets. SVRG only dominates the small-scale 22-dimensional dataset IJCNN, and SGD yields the best performance on other two datasets KDD04\_bio and 20newsgroups. SSGD is found to be sensitive about imbalanced data partition, such as MED11 and 20newsgroups, where the positive/negative data ratios are 1:25 and 1:20 respectively.

It is also important to investigate the generalization ability of the learned machine learning models. To this end, we also periodically record the objective values on the testing set. The results are displayed in Fig.~\ref{fig:1b}, from which it is easy to visually rule out the possibility that the generalization error and training objective value are inconsistent.

It is surprising that the standard SGD is among the best performers on nearly all of 9 datasets despite its simplicity. Based on the experiments we argue that the research of semi-stochastic algorithms shall investigate the balance of larger step size and increased iteration complexity, particulary in the era of large data and high dimension.


\begin{figure*}[t]
\begin{center}
\includegraphics[height=0.2\linewidth,width=0.32\linewidth,keepaspectratio=false]{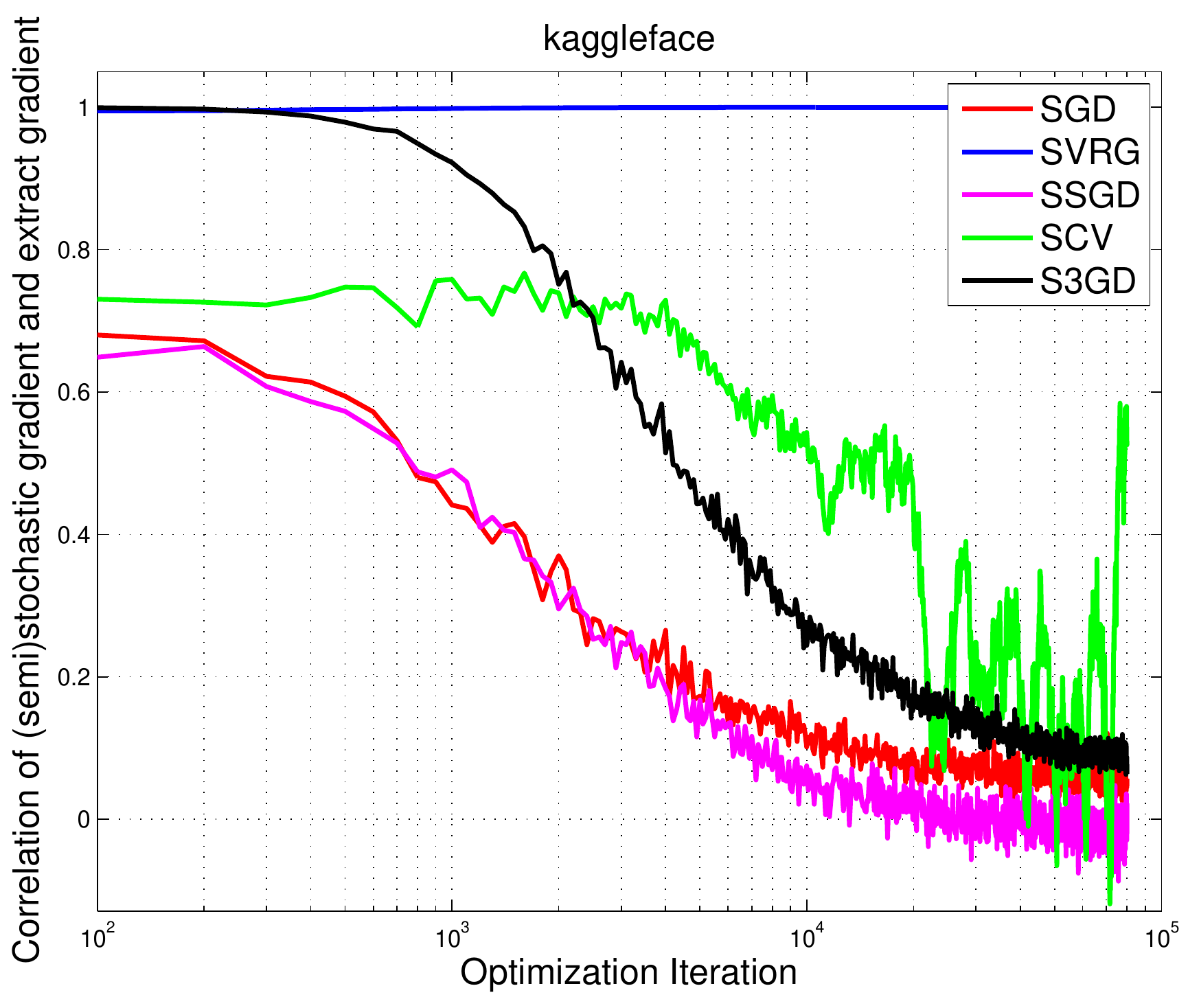}
\includegraphics[height=0.2\linewidth,width=0.32\linewidth,keepaspectratio=false]{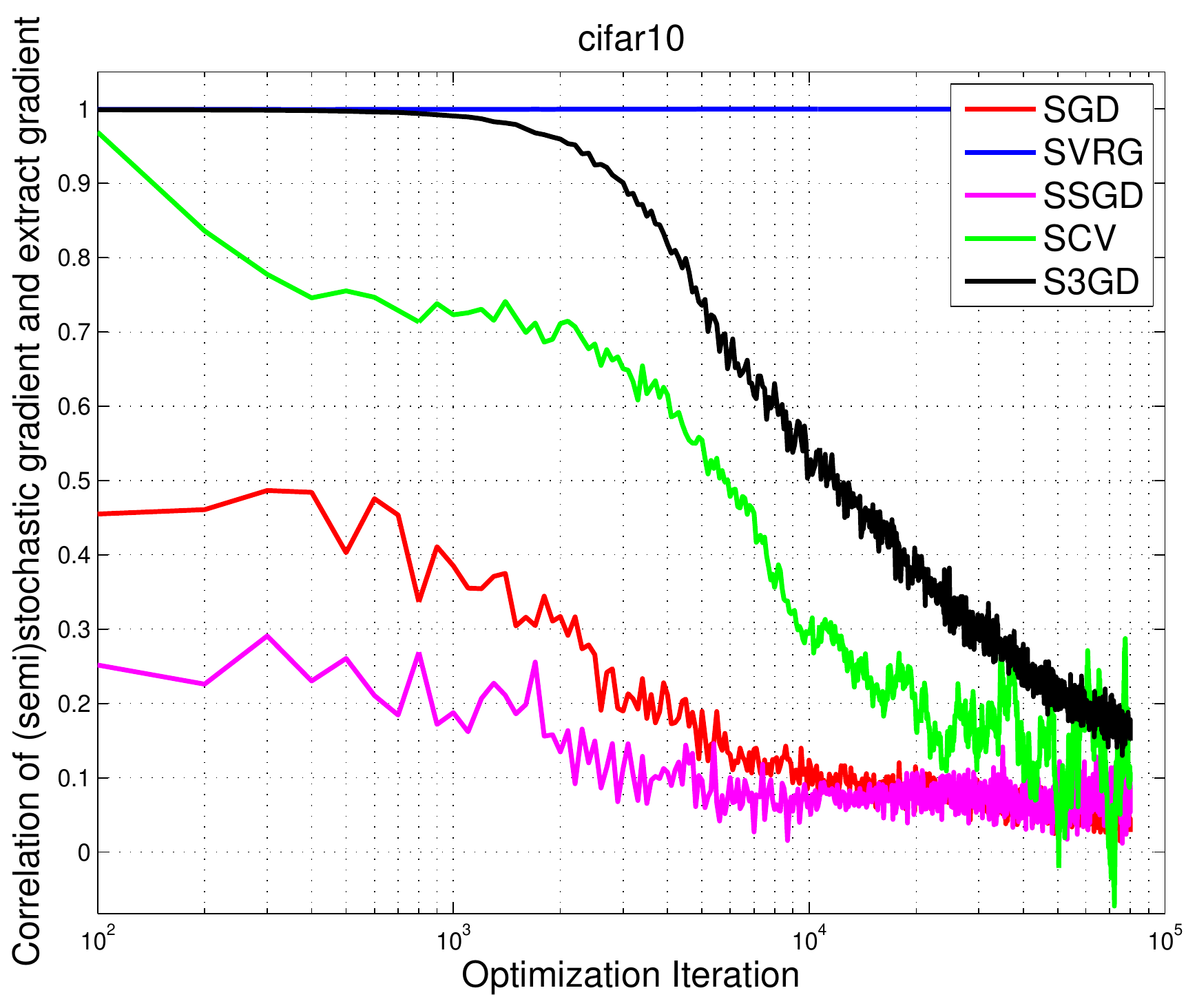}
\includegraphics[height=0.2\linewidth,width=0.32\linewidth,keepaspectratio=false]{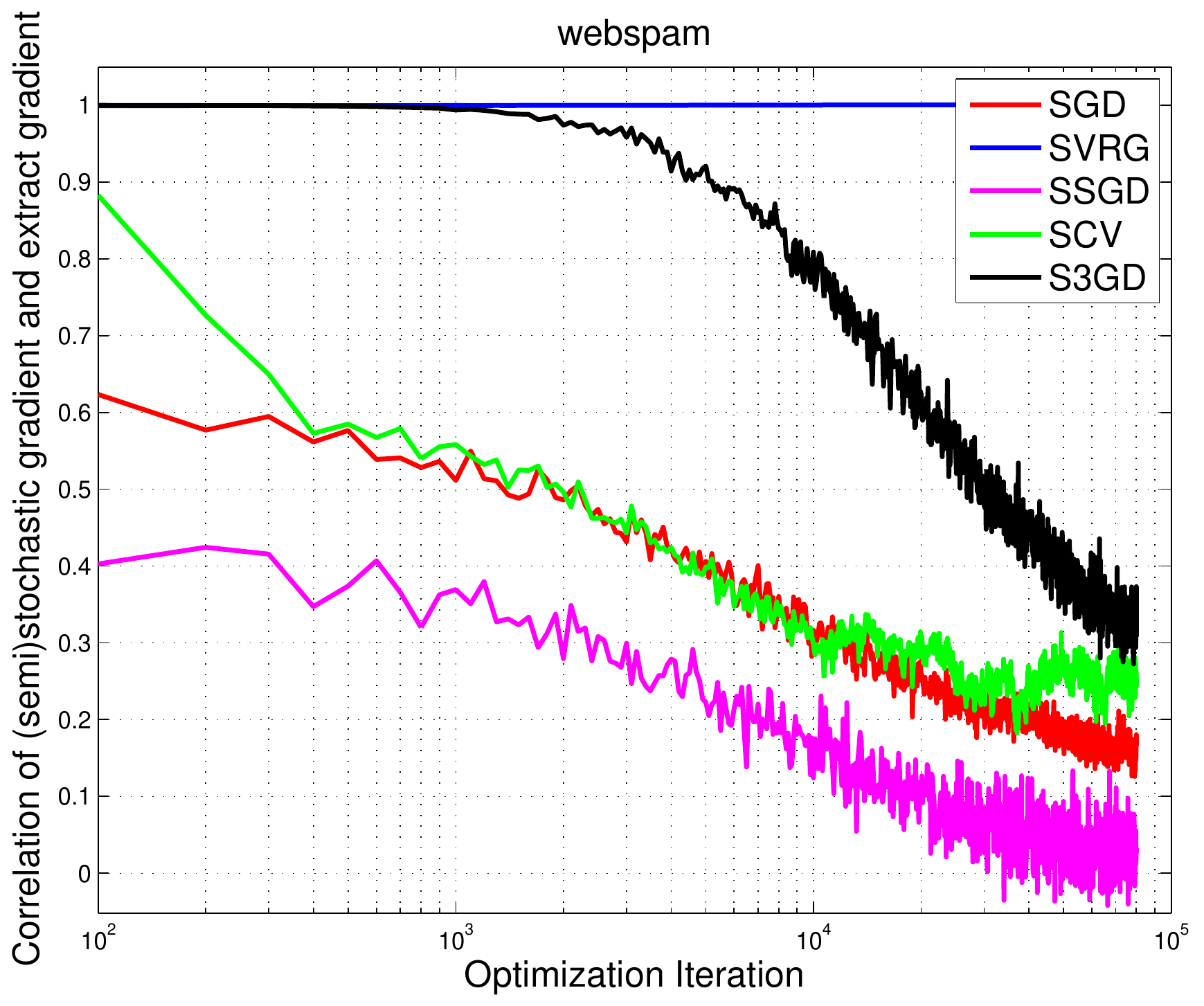}
\includegraphics[height=0.2\linewidth,width=0.32\linewidth,keepaspectratio=false]{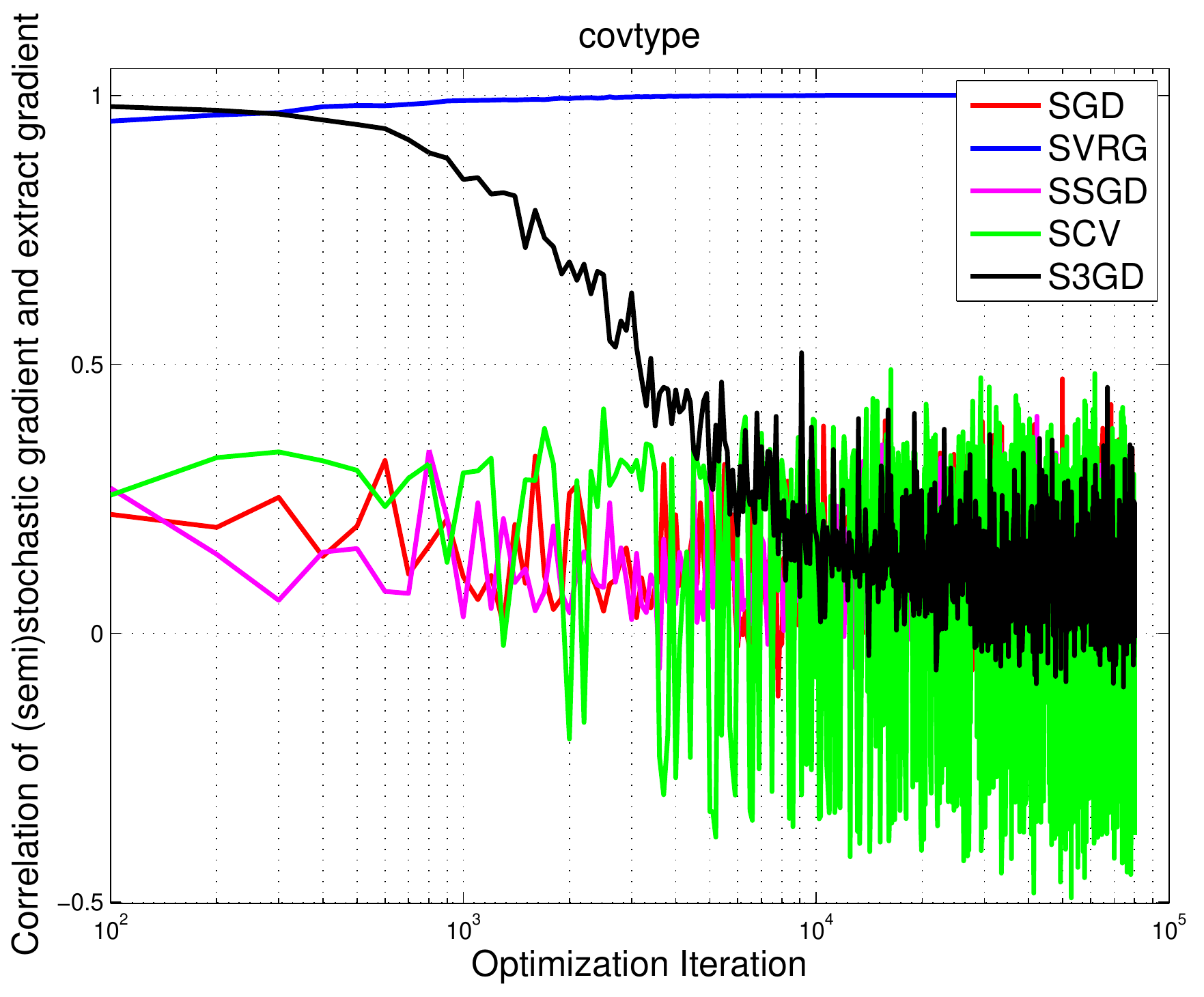}
\includegraphics[height=0.2\linewidth,width=0.32\linewidth,keepaspectratio=false]{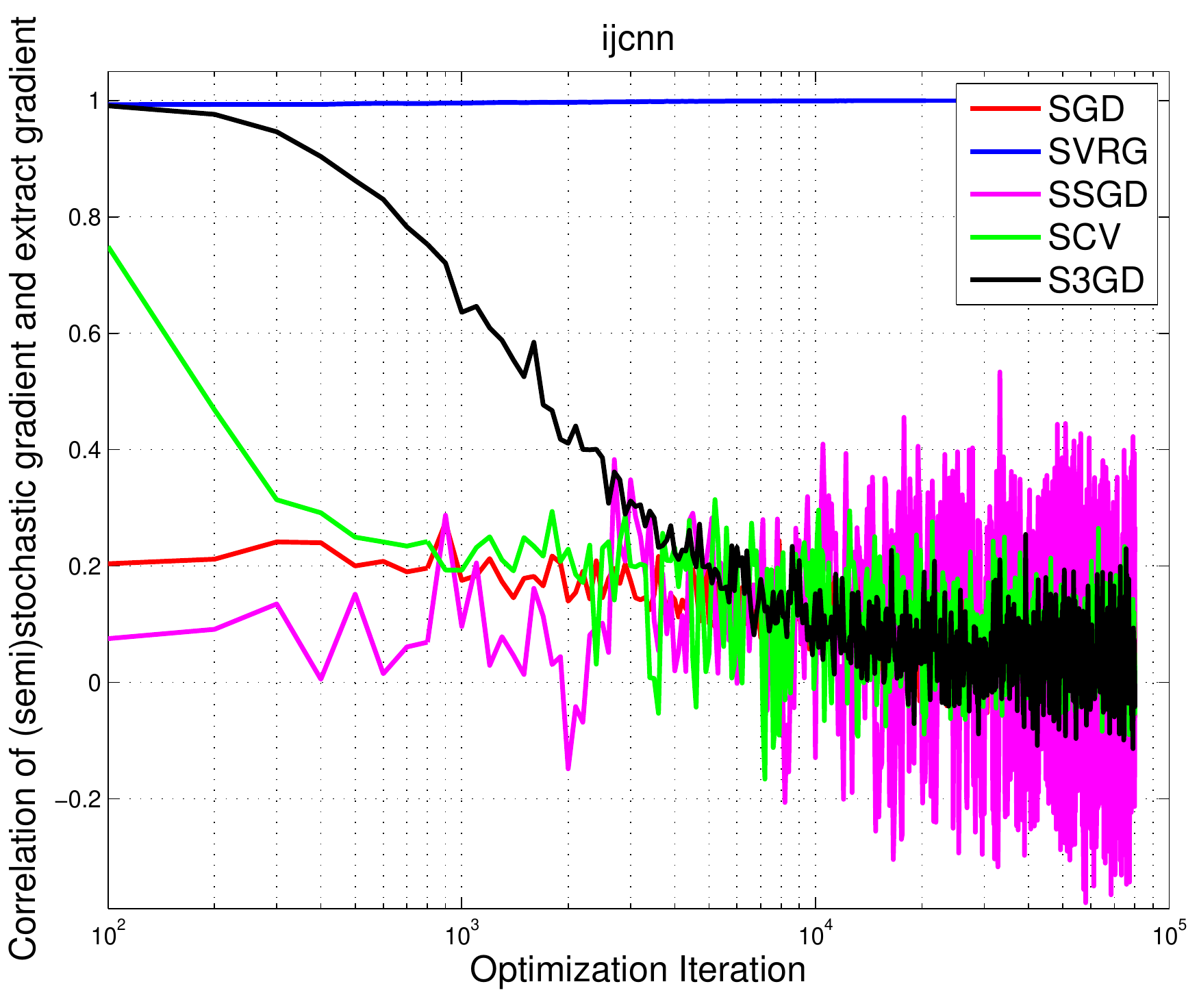}
\includegraphics[height=0.2\linewidth,width=0.32\linewidth,keepaspectratio=false]{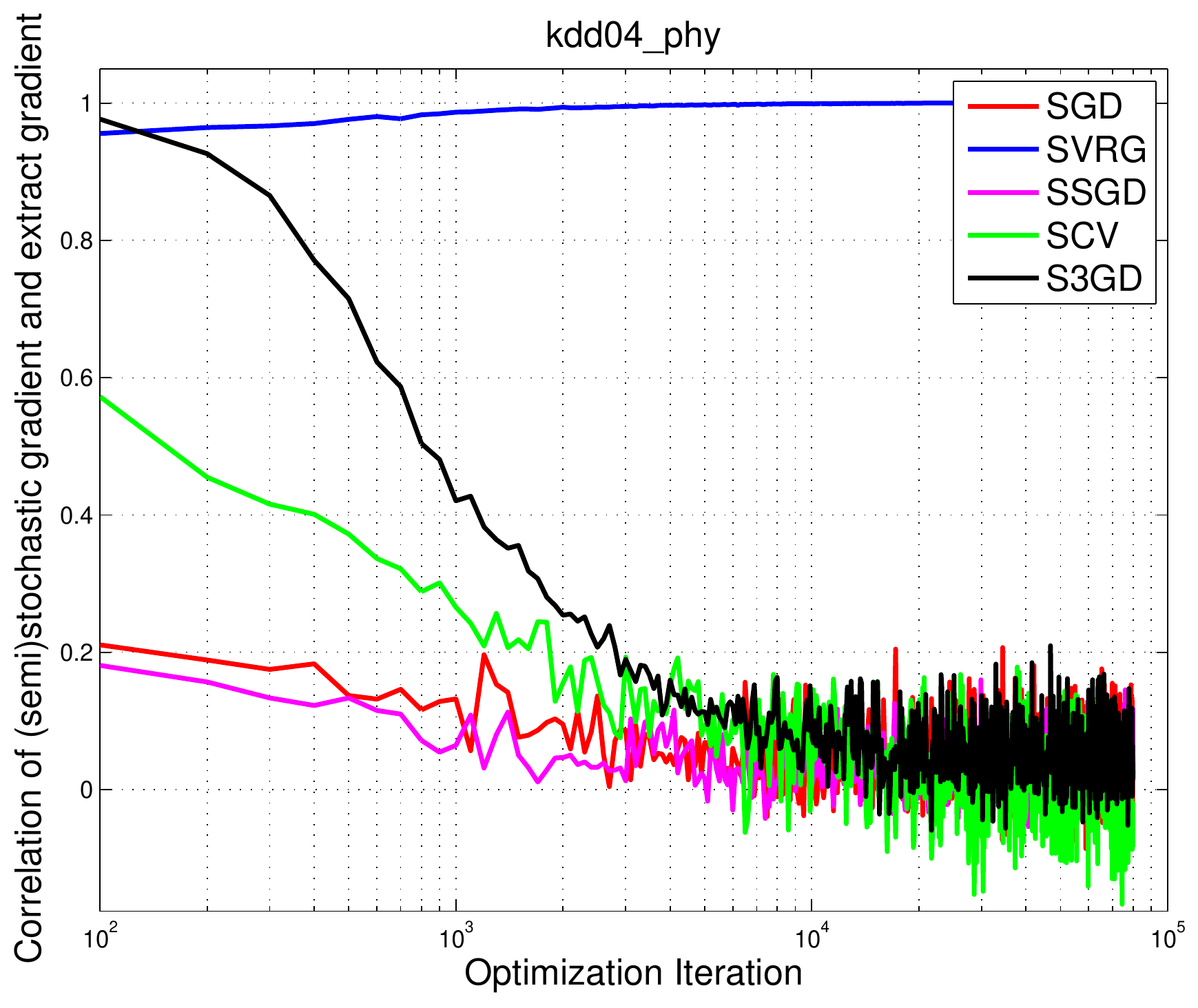}
\includegraphics[height=0.2\linewidth,width=0.32\linewidth,keepaspectratio=false]{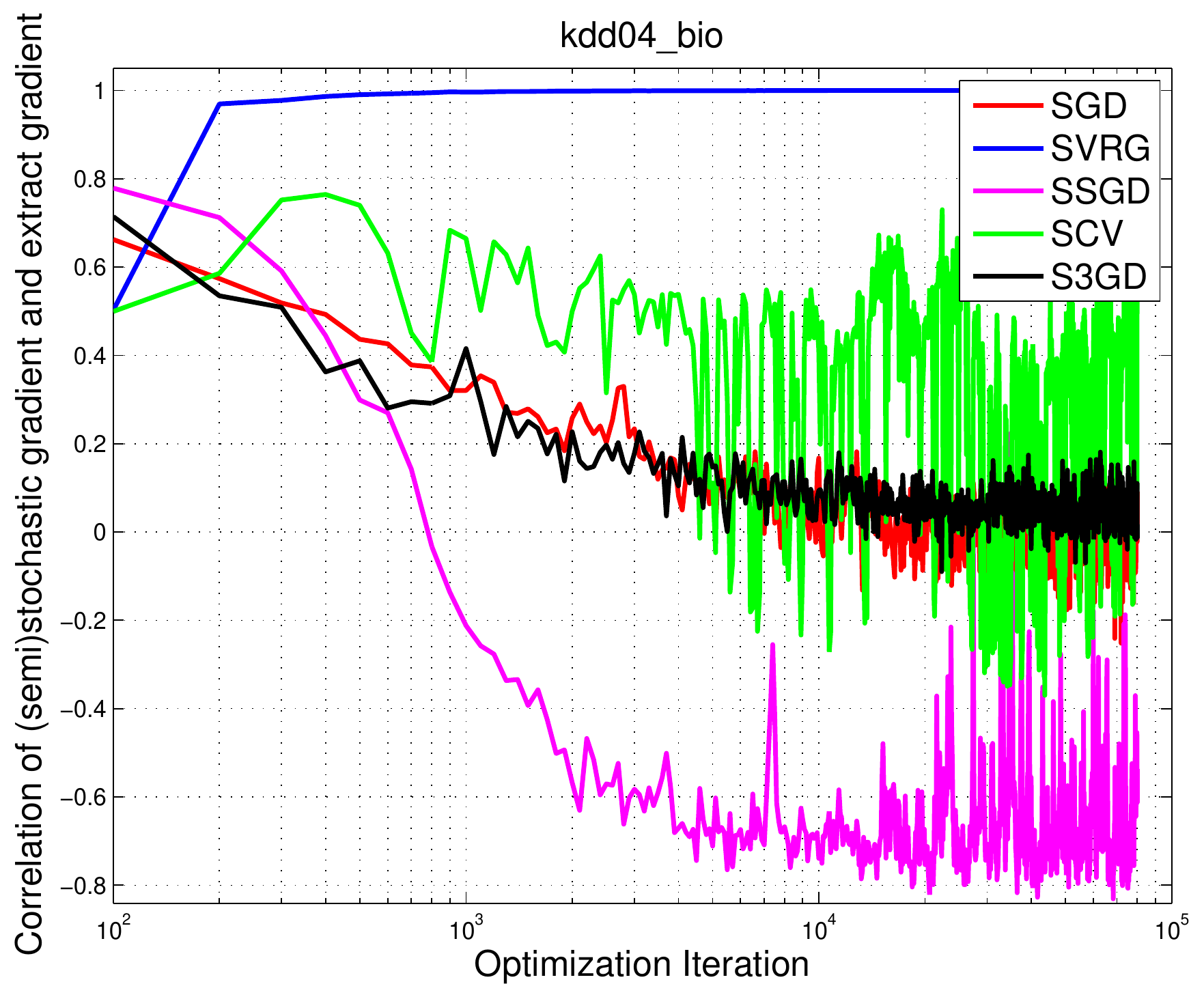}
\includegraphics[height=0.2\linewidth,width=0.32\linewidth,keepaspectratio=false]{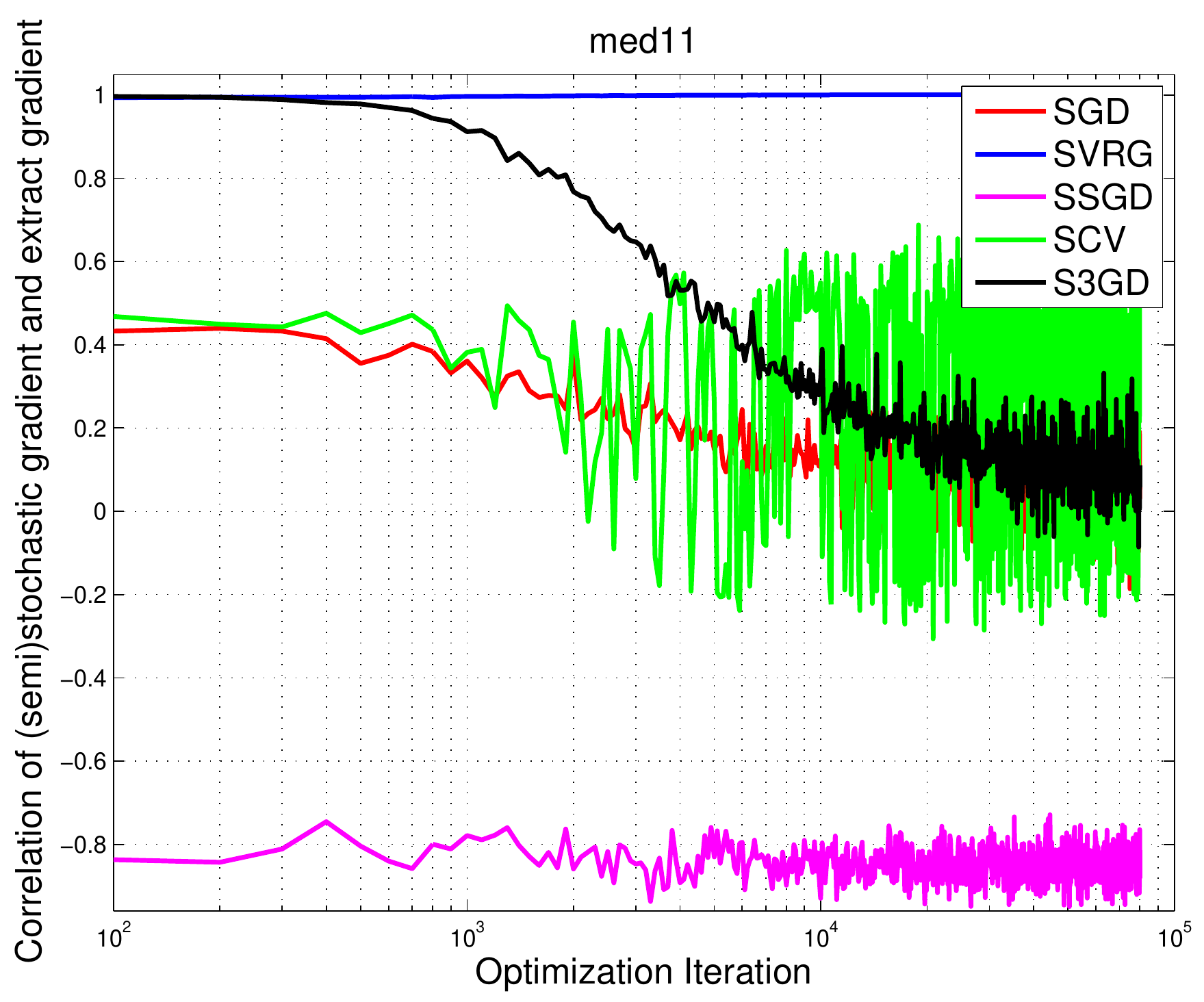}
\includegraphics[height=0.2\linewidth,width=0.32\linewidth,keepaspectratio=false]{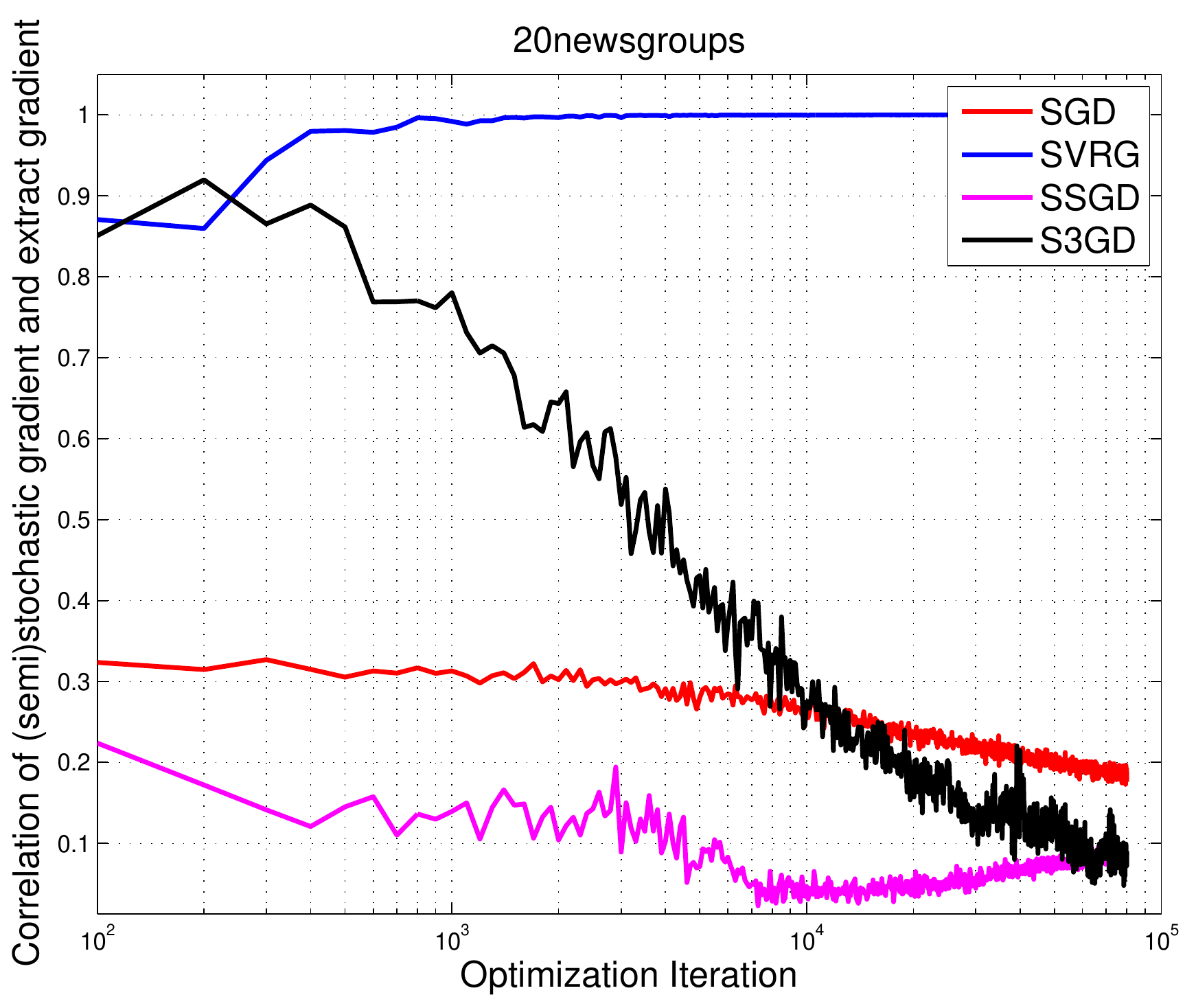}
\end{center}
   \caption{\small Pearson correlation scores of (semi)stochastic gradient and the exact gradient on 9 datasets under the parameter $\eta = 0.1$. See text for more explanation. Best viewing in color.}
\label{fig:2}
\end{figure*}

\vspace{0.07in}
\noindent \textbf{Corrleation of Gradients}: We further study the Pearson correlation of (semi)stochastic gradient and the exact gradient. For a semi-stochastic algorithm, the correlation score is favored to approach the value of 1, since it indicates a better approximation scheme for gradient computation.

It is clearly observed that SVRG and the proposed S3GD exhibit the most favorable correlation scores. Moreover, most methods enjoy relatively larger correlation scores when the optimization just begins. The correlation scores gradually drop when the optimization proceeds. The reason may be that the exact (sub)gradient tends to zero around the optimum, which makes accurate gradient approximation more challenging. The only exception is SVRG. In all cases its correlation scores quickly rise and stay at 1. It may be caused by the fact that $\| \w^{k+1} - \w^k \|$ tends to zero when approaching the global optimum. Therefore, $\widetilde \w \approx \w^k$ in Eqn.~(\ref{eqn:svrg}), which implies that the semi-stochastic gradient becomes increasingly close to the full gradient.

\begin{figure}[t]
\begin{center}
   \includegraphics[width=0.95\linewidth]{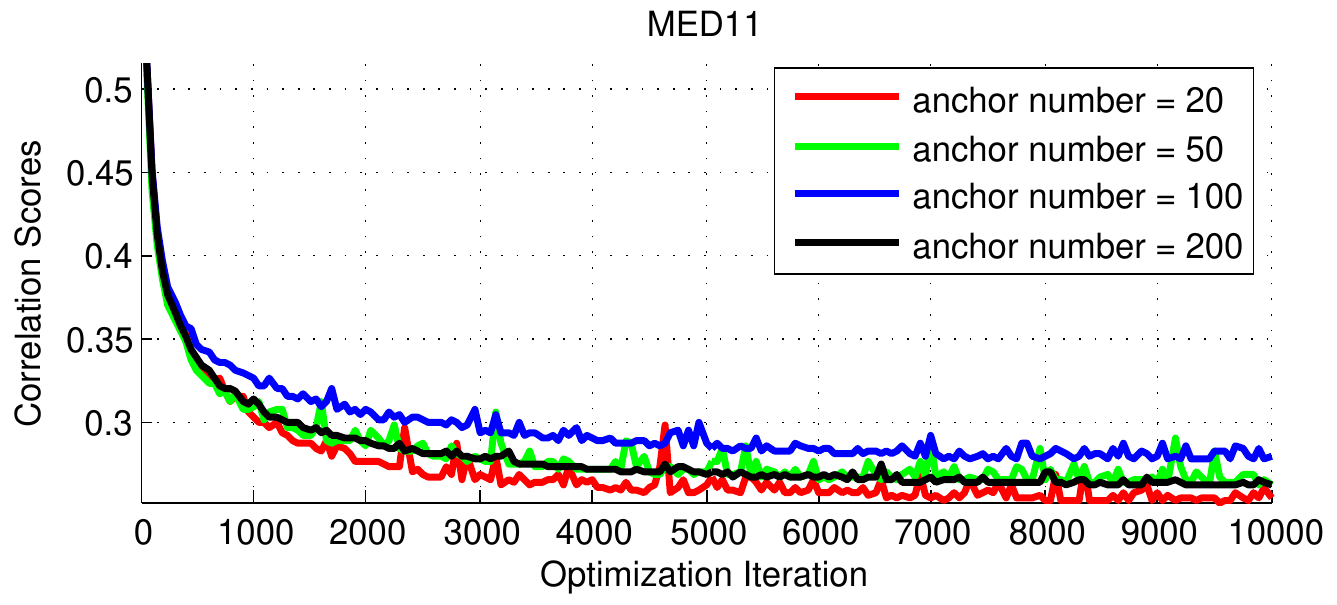}
   \includegraphics[width=0.95\linewidth]{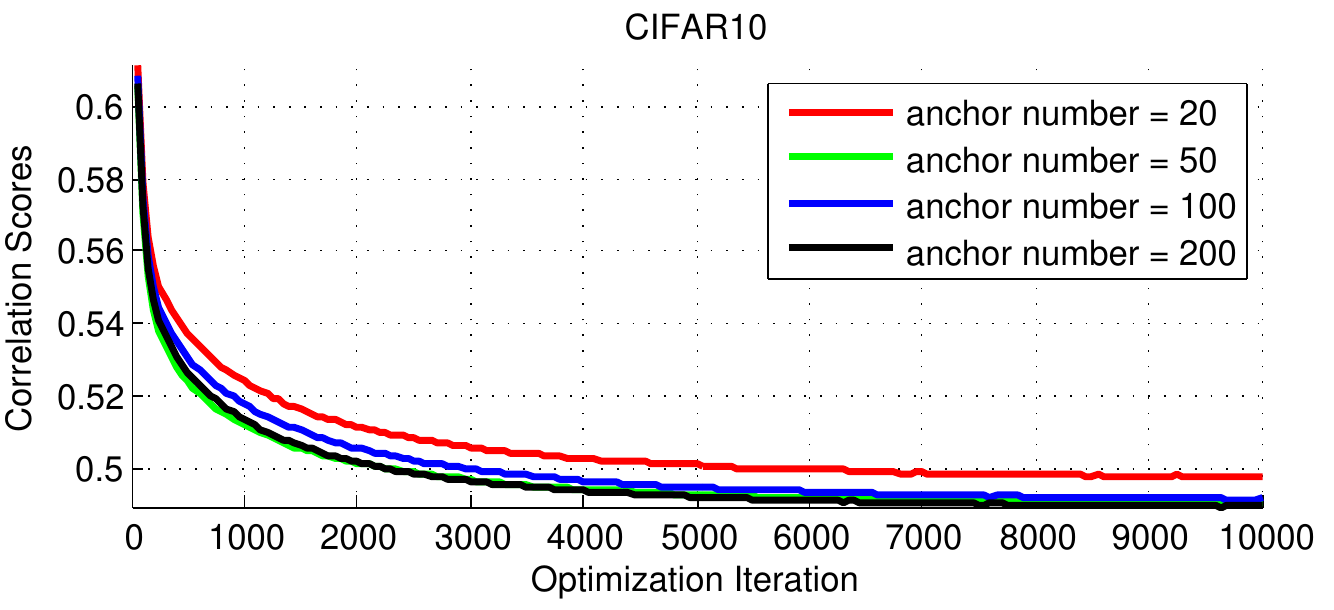}
\end{center}
   \caption{\small Investigation of how the anchors affect gradient correlation scores on MED11 and CIFAR10.}
\label{fig:3}
\end{figure}

\vspace{0.07in}
\noindent \textbf{Effect of Anchor}: Recall that we use 100 anchors obtained through clustering in all experiments. One may concern how different choices of anchor number affect the performance and running time. Figure~\ref{fig:3} presents the evolution of correlations scores under different anchor settings on MED11 and CIFAR10 (step size is fixed to be 1 for all cases).

\begin{figure}[t]
\begin{center}
   \includegraphics[width=0.8\linewidth]{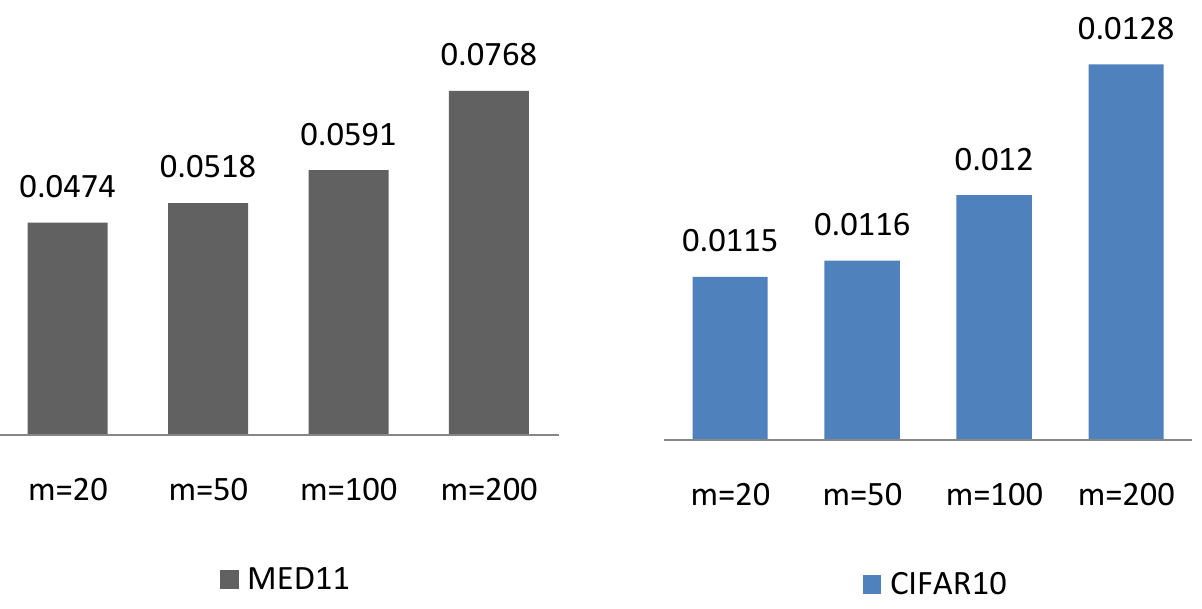}
\end{center}
   \caption{\small Average CPU time for every 50 optimization iterations (in seconds) for S3GD.}
\label{fig:4}
\end{figure}

Interestingly, we observe that enlarging anchor set does not entail boosted correlation scores. In fact, the scores will reach its peak around data-specific anchor number (100 for MED11 and 20 for CIFAR10) though other choices bring alike performances. This implies that the algorithm is largely robust to the anchor number though empirical tuning does further help. Figure~\ref{fig:4} plots the averaged iteration times for different anchor parameters. More anchors entail longer CPU time. Yet the time only increases sub-linearly owing to other kinds of computational overhead at each iteration.

\vspace{0.07in}
\noindent \textbf{Effect of inner iteration count $k_{in}$}: Both SVRG and our proposed S3GD adopt two-level loops for optimizing the objective function. The parameter $k_{in}$ refers to the maximal iteration count in the inner loops. Note that we use different $k_{in}$ for SVRG and S3GD (50 for SVRG and 20 for S3GD). Intuitively, smaller $k_{in}$ accelerates the convergence speed for both S3GD and SVRG with respect to the iterations, since it indicates more frequent update of the historical parameter vector at the outer loop. However, as shown in Table~\ref{table:complexity}, SVRG requires the computation of exact gradient at the beginning of each outer loop, which implies a complexity of $\mathcal{O}(n d)$. In contrast, S3GD only has a complexity of $\mathcal{O}(d m)$ using the pre-computation trick as in Eqn.~(\ref{eqn:HM}). Therefore, to balance the amortized iteration complexity and convergence speed, SVRG tends to prefer larger $k_{in}$ than S3GD.

\begin{figure}[t]
\begin{center}
   \includegraphics[width=0.95\linewidth]{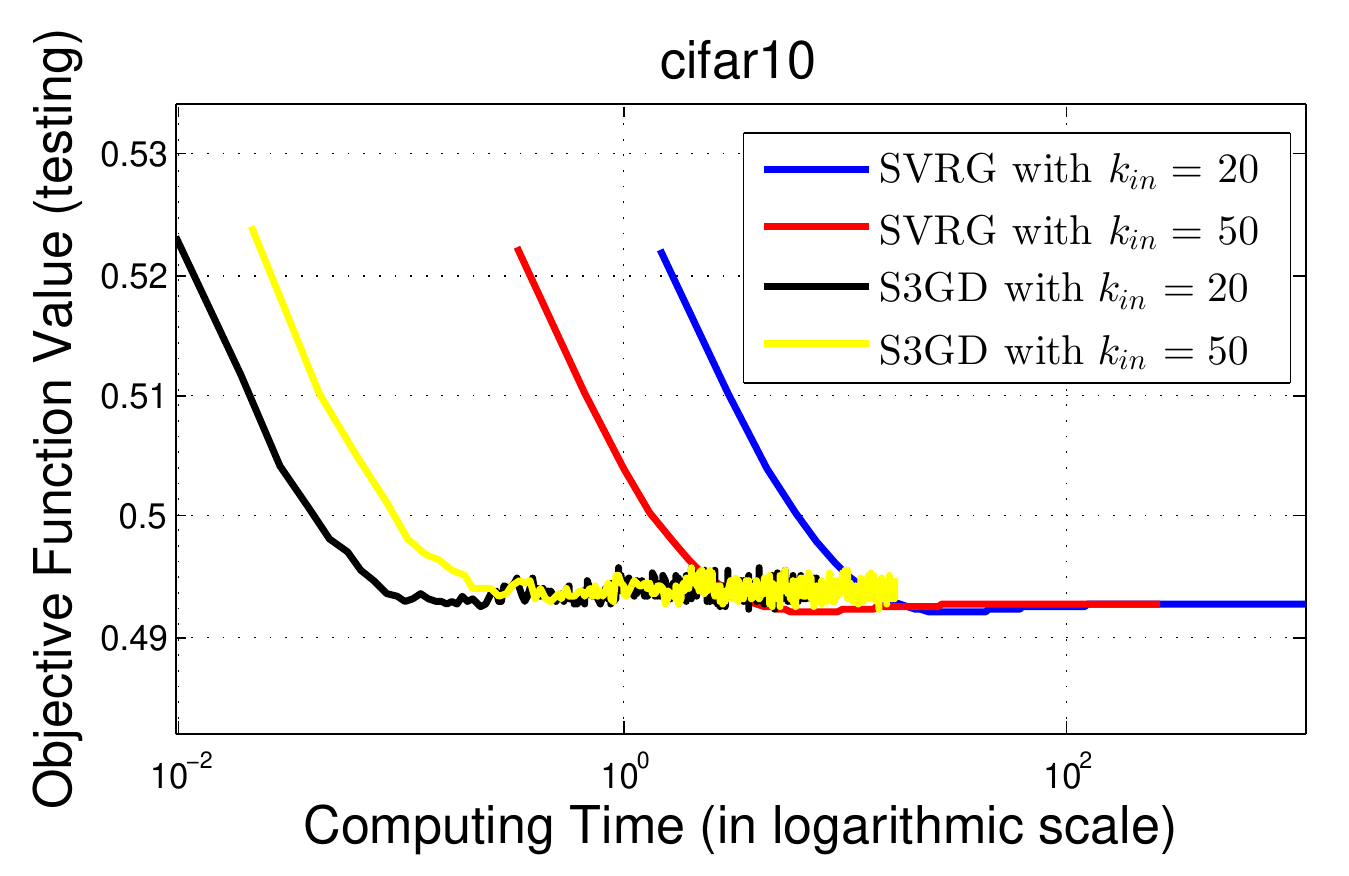}
\end{center}
   \caption{\small Investigation of how the parameter $k_{in}$ for maximal inner iteration count affects the convergence on CIFAR10.}
\label{fig:5}
\end{figure}

To validate this point, Fig.~\ref{fig:5} plots the objective values on the testing data of the CIFAR10 benchmark under $k_{in}=20$ or $k_{in}=50$. Both S3GD and SVRG exhibit similar convergence curves under different choices of $k_{in}$. Intuitively, though large $k_{in}$ implies more frequent historical parameter update, the significantly increased iteration complexity of SVRG actually slows down its convergence.


\section{Concluding Remarks}
\label{sec:conclusion}

In this paper we addressed the scalability issue pertaining to semi-stochastic gradient descent methods by proposing a novel approach S3GD. The motivation of S3GD is to reduce the high iteration complexity in existing semi-stochastic algorithms. It exploits stratified manifold-based gradient approximation as a good cure for the time-consuming exact gradient calculation. Our work significantly advances the original idea of residual-minimizing gradient correction. The current paper did not discuss the application in a distributed computing environment, since it is out of the main scope. However, we will explore the distributed variants of the proposed S3GD like~\cite{ShamirS014} in the future. Moreover, extension to non-convex formulations such as deep networks~\cite{deeplearning} is also a meaningful future direction.


%
%

\end{document}